\pdfoutput=1

\documentclass{article}

\usepackage{iclr2027_conference,times}

\usepackage{amsmath,amsfonts,bm}

\def\eqref#1{equation~\ref{#1}}

\def\1{\bm{1}}

\DeclareMathAlphabet{\mathsfit}{\encodingdefault}{\sfdefault}{m}{sl}
\SetMathAlphabet{\mathsfit}{bold}{\encodingdefault}{\sfdefault}{bx}{n}

\usepackage{graphicx}
\usepackage{wrapfig}
\usepackage[pagebackref]{hyperref}  
\usepackage{url}
\usepackage{fontawesome5}
\usepackage{booktabs}
\usepackage{tabularx}
\usepackage{placeins}  
\usepackage{float}     
\usepackage{multirow}
\usepackage{colortbl}   
\usepackage{capt-of}   

\usepackage[disable]{todonotes}
\usepackage[normalem]{ulem}   
\usepackage{xspace}
\usepackage[most]{tcolorbox}

\makeatletter
\newcommand*\iftodonotes{\if@todonotes@disabled\expandafter\@secondoftwo\else\expandafter\@firstoftwo\fi}
\makeatother

\definecolor{modtext}{HTML}{0F766E}   
\definecolor{modimage}{HTML}{7C3AED}  

\newcommand{\Mimg}{\textcolor{modimage}{img}}
\newcommand{\Mtxt}{\textcolor{modtext}{txt}}     
\newcommand{\qTT}{\Mtxt\ensuremath{\to}\Mtxt\xspace}
\newcommand{\qTI}{\Mtxt\ensuremath{\to}\Mimg\xspace}
\newcommand{\qIT}{\Mimg\ensuremath{\to}\Mtxt\xspace}
\newcommand{\qII}{\Mimg\ensuremath{\to}\Mimg\xspace}

\tcbset{
  inlinebox/.style={
    enhanced, breakable,
    colframe=#1!60!black, colback=#1!10,
    boxrule=0.6pt, arc=3pt,
    left=3pt, right=3pt, top=2pt, bottom=2pt,
    fontupper=\small
  }
}

\usepackage{cleveref}  
\crefformat{section}{\S#2#1#3}
\crefmultiformat{section}{\S\S#2#1#3}{ and~#2#1#3}{, #2#1#3}{, and~#2#1#3}
\crefrangeformat{section}{\S\S#3#1#4--#5#2#6}
\Crefformat{section}{\S#2#1#3}
\Crefmultiformat{section}{\S\S#2#1#3}{ and~#2#1#3}{, #2#1#3}{, and~#2#1#3}
\Crefrangeformat{section}{\S\S#3#1#4--#5#2#6}
\crefformat{subsection}{\S#2#1#3}
\crefmultiformat{subsection}{\S\S#2#1#3}{ and~#2#1#3}{, #2#1#3}{, and~#2#1#3}
\crefrangeformat{subsection}{\S\S#3#1#4--#5#2#6}
\Crefformat{subsection}{\S#2#1#3}
\Crefmultiformat{subsection}{\S\S#2#1#3}{ and~#2#1#3}{, #2#1#3}{, and~#2#1#3}
\Crefrangeformat{subsection}{\S\S#3#1#4--#5#2#6}
\usepackage[most]{tcolorbox}
\definecolor{slotcol}{HTML}{B45309}
\definecolor{poscol}{HTML}{1A7A1A}   
\definecolor{negcol}{HTML}{B02020}   
\newcommand{\slot}[1]{\textit{#1}}   
\definecolor{steercol}{HTML}{FFE270}   
\newcommand{\steer}[1]{{\setlength{\fboxsep}{1.2pt}\colorbox{steercol}{#1}}\kern-1.2pt}   
\definecolor{pillNeg}{HTML}{C0392B}\definecolor{pillNegL}{HTML}{E08A7E}
\definecolor{pillZero}{HTML}{8A8A8A}
\definecolor{pillPosL}{HTML}{7BB98A}\definecolor{pillPos}{HTML}{1E874B}
\newtcolorbox{samplebox}{enhanced, equal height group=txtsamples,
  colback=white, colframe=black!20, boxrule=0.7pt, arc=4pt,
  left=8pt, right=8pt, top=6pt, bottom=6pt}

\newtcolorbox{samplefull}{enhanced, width=\textwidth, colback=white, colframe=black!16, boxrule=0.6pt,
  arc=3pt, left=7pt, right=7pt, top=4pt, bottom=4pt}

\newtcolorbox{tsamplebox}[1][]{colback=black!3, colframe=black!45, boxrule=0.4pt, arc=3pt,   
  left=6pt, right=6pt, top=3pt, bottom=3pt, toptitle=1pt, bottomtitle=1pt, unbreakable, equal height group=txtqual,
  fonttitle=\bfseries\scriptsize, coltitle=black, colbacktitle=black!8, title={#1}}

\newcommand{\imtag}[1]{\textbf{\textless\textbar #1\textbar\textgreater}}   
\newcolumntype{Y}{>{\raggedright\arraybackslash\ttfamily\fontsize{5}{5.9}\selectfont}X}
\newcommand{\kt}[1]{\leavevmode{\bfseries #1}}                     
\newcommand{\ktb}[1]{\leavevmode{\color{black!55}#1}}              
\newcommand{\ka}[2]{{\fontsize{5.5}{6.5}\selectfont\bfseries\color{#1}#2}}
\newcommand{\kp}[2]{\leavevmode{\rmfamily\fontsize{6.5}{7.5}\selectfont\bfseries\color{pillNeg}#1}\,/\,{\rmfamily\fontsize{6.5}{7.5}\selectfont\bfseries\color{pillPos}#2}}   
\newcommand{\kms}[3]{\begin{tabular}[t]{@{}l@{}}\ka{#1}{#2}\\[-1pt]{\fontsize{4.6}{5.2}\selectfont\color{black!55}#3}\end{tabular}}   
\newcommand{\ktbox}[1]{\begin{tcolorbox}[colback=black!3, colframe=black!45, boxrule=0.4pt, arc=3pt, left=3pt, right=3pt, top=1.5pt, bottom=1.5pt, boxsep=0pt]
\setlength{\tabcolsep}{2.5pt}\renewcommand{\arraystretch}{1.0}\fontsize{5}{5.9}\selectfont #1\end{tcolorbox}}
\newcommand{\vtag}[1]{\textbf{\textless #1\textgreater}}                     
\newcommand{\promptlegend}{\\[7pt]{\rmfamily\scriptsize\color{black!60}\textcolor{steercol}{\rule[-0.1ex]{1.3em}{1.2ex}}~steered tokens\quad\textbf{bold}~chat template}}
\newtcolorbox{cpromptbox}[1][]{%
  colback=black!2, colframe=black!45, boxrule=0.4pt, arc=3pt,
  left=9pt, right=9pt, top=9pt, bottom=9pt, toptitle=2pt, bottomtitle=2pt, unbreakable,
  fonttitle=\bfseries\footnotesize, coltitle=black,
  colbacktitle=black!7, title={#1}}
\newtcolorbox{jpromptbox}[1][]{
  colback=black!2, colframe=black!45, boxrule=0.4pt, arc=3pt,
  left=9pt, right=9pt, top=9pt, bottom=9pt, toptitle=2pt, bottomtitle=2pt, unbreakable,
  fonttitle=\bfseries\footnotesize, coltitle=black,
  colbacktitle=black!7, title={#1}}
\newtcblisting{promptbox}[1][]{
  listing only,
  listing options={basicstyle=\ttfamily\footnotesize,breaklines=true,breakatwhitespace=true,columns=fullflexible},
  colback=black!2, colframe=black!45, boxrule=0.4pt, arc=3pt,
  left=9pt, right=9pt, top=9pt, bottom=9pt, toptitle=2pt, bottomtitle=2pt, unbreakable,
  fonttitle=\bfseries\footnotesize, coltitle=black,
  colbacktitle=black!7, title={#1}
}

\title{Cross-modal Steering Vectors Exist Across Both Image and Text Generation
}

\newcommand{\aff}[1]{\textsuperscript{#1}}
\author{%
Donghao Zeng\aff{1} \quad Nishant Subramani\aff{2} \quad Tamar Rott Shaham\aff{3} \quad Benno Krojer\aff{1,4}\thanks{Work done primarily at Mila, McGill University. Correspondence to \texttt{bkrojer@northeastern.edu}.} \\
\aff{1}Mila, McGill University \quad \aff{2}Carnegie Mellon University, LTI \\
\aff{3}MIT CSAIL \quad \aff{4}Northeastern University
}

\iclrfinalcopy

\begin{document}

\maketitle
\ificlrfinal\lhead{Preprint. Under review.}\fi

\begin{abstract}
Are concepts and mechanisms unified across modalities in current AI models, or instead implemented separately for each one?
We operationalize this broad question of modality unification by studying \textit{cross-modal steering}:
1) finding vectors that successfully steer generation on tokens of one modality, which then 2) transfer zero-shot to steer generation when added to tokens of another modality.
Our study goes beyond simple correlational alignment (e.g. representational similarity) because steering is fundamentally causal. 
We study cross-modal steering in several unified multimodal models (UMMs), since they possess shared parameters for language and image generation, focusing on these two modalities.
Across diverse steered concepts such as emotion and object size, we show that steering vectors transfer zero-shot in both directions (text-to-image \& image-to-text).
Even for concepts that are highly unique to one modality (e.g. formal vs. colloquial language), steering vectors often induce a consistent, interpretable effect in the other modality, such as formal clothing and design.
This transfer has a geometric signature too: despite a clear modality gap separating text and image representations, steering vectors with strong cross-modal transferability tend to point in similar directions for the same concept.
Our findings shed light on how vision and language interact inside unified multimodal models and point towards more general, modality-agnostic approaches for controlling their behavior.

\vspace{0.5em}
\begin{center}
\ificlrfinal
\faSearch\ \href{https://saintsauceee.github.io/unified-cm-steer-demo}{\textsc{Demo}}
\quad
\faGithub\ \href{https://github.com/saintsauceee/cross-modal-steering}{Code}
\else
\fi
\end{center}
\end{abstract}

\section{Introduction}

Despite the complexity of modern generative models, such as LLMs and diffusion models, many high-level concepts can be controlled through surprisingly simple interventions: adding a linear direction, or \emph{steering vector}~\citep{subramani-etal-2022-extracting}, to a model's intermediate representations. Steering vectors have been used to modulate a wide range of behaviors in both language models~\citep{li2023inferencetimeintervention,todd2024functionvectors, turner2024steeringlanguagemodelsactivation,dunefsky2025oneshot,bayat2025steering} and image generation systems~\citep{GanDissect,goetschalckx2019ganalyze,spingarn2021gan,park2023unsupervisedsemanticlatent,gandikota2024conceptsliders, konovalova2026shiftsteeringhiddenintermediates}, providing a lightweight and causal means of controlling model outputs. Yet steering in these two modalities has largely been studied separately. Here we ask: are steering directions themselves modality-specific, or can a direction discovered in one modality also control generation in another?

\begin{figure}[t]\centering
\includegraphics[width=\textwidth]{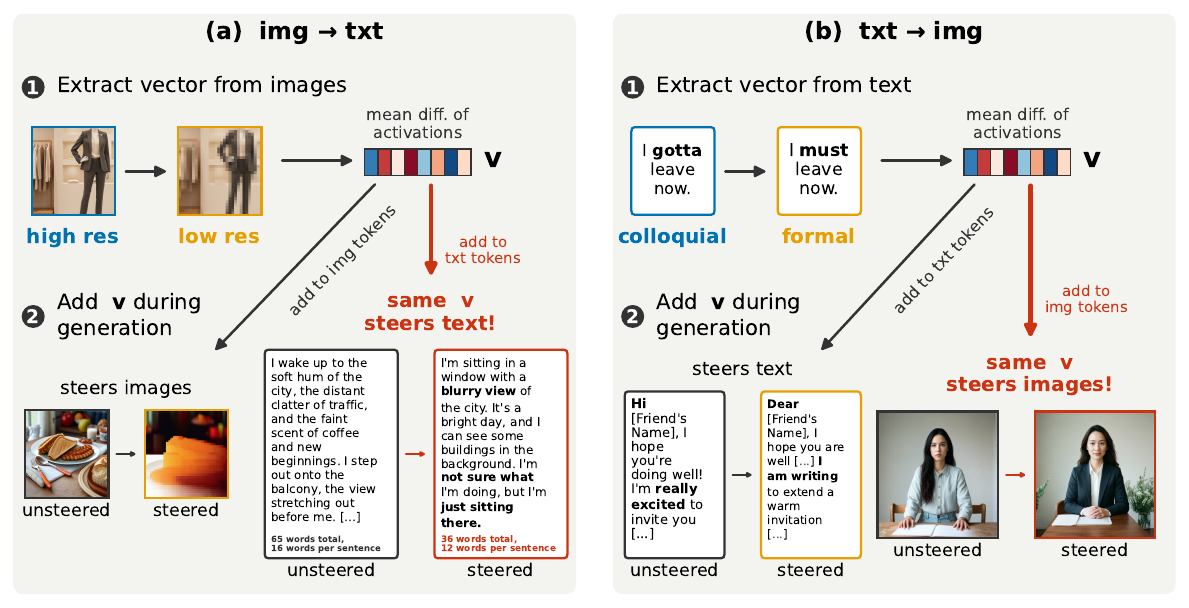}
\vspace{-6mm}
\caption{\textbf{Cross-modal steering vectors transfer zero-shot in both directions.} 
A steering vector $\mathbf{v}$ extracted entirely from one modality can steer generation both within that modality and, without any adaptation, in the other.
%
}
\label{fig:teaser}
\label{fig:retention-radar}
\end{figure}

Consider a steering vector extracted entirely from text that, when added to a model's hidden states, reliably causes it to write more formally. If the same model also generates images, what happens when that very same vector is instead added to its generated image tokens? Could it induce a corresponding notion of \textit{visual formality}, such as more formal clothing or design? And vice versa: could a direction extracted purely from images steer how the model writes? 
We call this \textit{cross-modal steering}, illustrated in \Cref{fig:teaser}, and ask:
\textbf{Can a steering vector constructed only on tokens from one modality generalize zero-shot when added to tokens from the other modality?}

This question connects to a broader debate about how concepts are represented across modalities. On the one hand, recent evidence from multimodal models suggests substantial internal separation between language and vision: text and image token representations remain nearly orthogonal throughout the layers \citep{cite_narrow_gate},
knowledge edited on the text side mostly fails to show up in generated images \citep{gao2026text}, and
circuits in multi-modal LLMs for the same task are often disjoint \citep{nikankin2025same}. 
On the other hand, cognitive science has long proposed a modality-agnostic ``semantic hub'' where concepts live regardless of whether they arrive through language or perception \citep{patterson2007semantichub}.
We investigate this through cross-modal steering: successful transfer would demonstrate that a common direction causally modulates a concept across modalities, even when their representations may occupy distinct regions in activation space.

In the mainstream paradigms of multimodal modeling \citep{rombach2022ldm,liu2023llava, wang2024qwen2vl}, it would not be possible to answer this question since image-to-text and text-to-image generation are separately parameterized and modeled either via autoregression \citep{radford2019language} or diffusion \citep{sohldickstein2015diffusion}.
Thus, we turn to unified multimodal models (UMMs) as a natural testbed for cross-modal steering \citep{wang2024emu3nexttokenpredictionneed,chameleonteam2025chameleonmixedmodalearlyfusionfoundation, zhou2024transfusionpredicttokendiffuse, chen2025janusprounifiedmultimodalunderstanding, jiao2025unitokenharmonizingmultimodalunderstanding, xie2025showo}.
Some frontier proprietary systems have similarly moved to generating images natively within a shared language-model architecture rather than calling a separate image model \citep{geminiteam2023gemini, openai2025nativeimagegeneration}. Yet shared parameters do not imply shared internal mechanisms, making cross-modal transfer far from guaranteed. Indeed, prior work studying cross-modal steering in UMMs reports strong asymmetry: text-derived vectors transfer well to image generation, but image-derived vectors struggle to transfer to text generation including limited transfer in autoregressive models~\citep{wang2026unifiedmultimodalmodelsthink}.



\textbf{We find that steering vectors transfer across modalities in both directions.} We tested three UMMs that autoregressively generate text and image tokens from a shared weight space: UniAR \citep{peng2026unified}, Emu3.5 \citep{cui2025emu35nativemultimodalmodels} and Liquid \citep{wu2026liquid}.
In UniAR
, the transferred vector is almost as good as the intra-modality vector for steering, recovering 84\% of its steering effect on images and 74\% on text.
In Emu3.5 and Liquid, text-derived vectors sometimes even steer images better than image-derived ones.
This illustrates the promise of studying cross-modal steering: a steering vector can be constructed in whichever modality is easier, and then used in the other.
This matters most for concepts that are highly unique to one modality (see \Cref{fig:teaser}).
How such a vector would express itself in the other modality gives a glimpse of how the model bridges vision and language:
steering an image generator with a vector for formal vs.\ colloquial \textit{language} causes people or rooms to be depicted more formally, and steering a language model with a vector for low image \textit{resolution} causes the model to write shorter text and express uncertainty.
Finally, we show that cross-modal steering has a geometric signature: despite a clear modality gap between text and image representations, the text and image vectors of many concepts point in similar directions. The closer they are, the better they facilitate transfer (\Cref{sec:geometry}).

Together, our results show that modality-specific representations can nevertheless contain shared causal directions of control, revealing a form of internal structure that is neither fully modality-specific nor captured simply by representational alignment. Beyond offering a new lens on how vision and language interact inside unified models, cross-modal steering points toward modality-agnostic approaches for controlling multimodal generation.

\section{Cross-Modal Steering: Methodology}
\label{sec:method}

\subsection{Preliminaries}

\paragraph{Unified Multimodal Models.}
In our study, we consider unified multimodal models (UMMs) that generate both language and visual content in one autoregressive stream, from an underlying shared weight space \citep{wang2024emu3nexttokenpredictionneed, chameleonteam2025chameleonmixedmodalearlyfusionfoundation,   cui2025emu35nativemultimodalmodels,wu2026liquid, peng2026unified, zhao2026unifiedmultimodalunderstandinggeneration}.
Concretely, a UMM models a mixed sequence $x = (x_1, \dots, x_T)$, in which every token $x_t$ is either a text token or a discrete visual token, with a single next-token objective:
\[
p_\theta(x) = \prod_{t=1}^{T} p_\theta\!\left(x_t \mid x_{<t}\right),
\]
where the parameters $\theta$ are shared across modalities,
The term \textit{unified} is used broadly in the literature, and can even include models with mostly separate parameters and mixed objectives \citep{deng2025emergingpropertiesunifiedmultimodal}.
For our study, we require models to share the weights responsible for generation.
A shared autoregressive objective additionally lets us extract and apply steering vectors in the same manner on both token types.
We study three such UMMs: UniAR (8B) \citep{peng2026unified}, Emu3.5 (34B) \citep{cui2025emu35nativemultimodalmodels} and Liquid (7B) \citep{wu2026liquid}.
They differ most in how images are tokenized: UniAR's image tokens are semantic, derived from a language-aligned vision encoder, whereas those of Emu3.5 and Liquid are trained to reconstruct pixels.

\paragraph{Steering Vectors.}
Steering is any intervention on model internals that allows continuous modulation of the output generation or a behavior by adding a vector to the intermediate hidden states during generation~\citep{subramani2019can, subramani-etal-2022-extracting, li2023inferencetimeintervention, turner2024steeringlanguagemodelsactivation}:
\[
h_l' = h_l + \alpha\, v_l.
\]
Here $h_l$ is a hidden representation at layer $l$, $v_l$ is the steering vector, and the scalar $\alpha$ sets the strength and sign of the intervention.
The two key design choices are a) how to find such a vector and b) how to apply it during generation.
In LLMs, steering vectors are typically found from contrastive prompt pairs or from probes and sparse dictionaries trained on the residual stream, and applied to one or many layers at every step of token-by-token decoding \citep{li2023inferencetimeintervention, rimsky2024contrastiveactivationaddition, huben2024sparseautoencoders}.
In image generation, directions are instead often discovered in the latent space of a diffusion model (e.g., by unsupervised decomposition) or learned as lightweight adapters, and applied along the denoising trajectory \citep{kwon2023diffusionsemanticlatent, park2023unsupervisedsemanticlatent, haas2024discoveringinterpretabledirectionssemantic, gandikota2024conceptsliders}.
The contrastive recipe that is standard for LLMs is less common in vision \citep{haas2024discoveringinterpretabledirectionssemantic, gaintseva2025casteer, konovalova2026shiftsteeringhiddenintermediates}, in part because minimal image pairs are much harder to curate than minimal text pairs.

\subsection{Our approach}
\label{sec:approach}
We adopt one established approach for both language and vision: we find steering vectors via a dataset of minimally different inputs, and add them to representations at either a subset of layers or whole ranges of layers. Crucially, finding and applying the vector need not use the same modality: for cross-modal steering, we construct a vector entirely from examples in one modality and then apply that \emph{same vector, without any adaptation or target-modality data}, to representations of the other modality during generation.
This gives four steering settings, corresponding to all combinations of finding a vector in either modality and applying it to either modality, as illustrated in \Cref{fig:teaser}.
Throughout the paper, we refer to them as \qTT, \qTI, \qIT, and \qII: \qTI and \qIT are the transfer settings we primarily study, and the intra-modality settings \qTT and \qII serve as reference points.
We color source and target modalities throughout (\Mtxt, \Mimg).



\paragraph{Extracting steering vectors.}
\label{sec:finding-vectors}

We extract steering vectors in a very similar way for text and images.
Given a concept and its associated poles (e.g., emotion: happy/sad), we generate $N=100$ minimal pairs $(x_i^{+}, x_i^{-})$ that differ only in the target pole.
We pass both inputs through the model and extract one representation per input at each layer $\ell$.
The steering vector is then computed as the average difference between paired representations:
\[
v_\ell = \frac{1}{N}\sum_{i=1}^N\bigl(h_{i,\ell}^{+}-h_{i,\ell}^{-}\bigr).
\]
We normalize each layer's steering vector to unit length, $\hat{v}_\ell = v_\ell / \lVert v_\ell \rVert$, so that the steering strength is set by a single coefficient that is comparable across layers and concepts.
The two modalities differ only in how the pairs are built and how an input is reduced to a single representation $h$.

\textit{Text.}
We automatically generate the pairs with Gemini 2.5 Flash \citep{geminiteam2025gemini25} that produces a diverse list of subjects that we fill into a fixed contrastive template.
Each pair consists of two short declarative prompts that only differ in the target attribute, which is placed as the last token, e.g., \textit{a person is happy} and \textit{a person is sad}.
We use the hidden state of the last token as $h$.

\textit{Image.}
We first generate a set of baseline images with Gemini 2.5 Flash Image.
We then edit each baseline image towards the two opposite values of the target attribute with an instruction-based image-editing model, for example \textit{make the person happier} and \textit{make the person sadder}, while mentioning in the prompt to keep the same subject, pose, framing, and background.
For most concepts we edit with the same model. For size, depth, and spatial, where its edits are less reliable, we use Gemini 3 Pro Image and Reve\footnote{\url{https://blog.reve.com/posts/the-reve-api/}}.
For simplicity, we mean-pool the hidden states across all $M$ image tokens, $h = \frac{1}{M}\sum_{j=1}^{M} h_j$.

\paragraph{Applying steering vectors.}
\label{sec:applying-vectors}

Given a steering vector, we intervene on the hidden states of all generated tokens, and for text, also on those of the subject phrase in the prompt, in the shared transformer backbone: at layer $\ell$, we set $h_\ell' = h_\ell + \alpha\, \hat{v}_\ell$, where $\alpha$ controls the steering strength and $\hat{v}_\ell$ is the per-layer unit steering vector defined above.
The intervention is identical whether the generated tokens are text or image tokens, and whether $\hat{v}_\ell$ was extracted from text or from images; in the transfer settings, the vector is simply added to tokens of the other modality.
We initially explore several strategies for \emph{where} in the model to apply this steering (single layers, contiguous ranges of layers, or all layers at once).
We focus on ranges of layers in \Cref{sec:results} --- early, in the middle, or late in the model --- since they steer more strongly than single layers and, unlike steering all layers at once, show \emph{where} in the model the two modalities align\footnote{\Cref{app:full-sweep} also shows results for single-layer and all-layer settings.}.
We sweep the steering strength $\alpha$ over a range chosen per model and setting, because usable strengths differ greatly between text and image generation and are larger for transferred vectors.
Each range extends just past the point where outputs break down (incoherent text, broken images).

\paragraph{Implementation details.}
\label{sec:experimental-details}
\textit{Concepts:} We study eight concepts: four \emph{semantic} concepts that naturally occur in both modalities, emotion (sad/happy), age (young/old), cleanness (clean/dirty) and chaos (peaceful/chaotic), and four \emph{visual} concepts that are spatial, geometric or appearance properties of a scene, size (small/large), depth (far/near), spatial (left/right) and color (blue/red).
We choose concepts that are commonly steered or studied in language models \citep{turner2024steeringlanguagemodelsactivation, zou2025representationengineeringtopdownapproach} or in image generation \citep{shen2020interfacegan, gandikota2024conceptsliders, baumann2025continuoussubjectspecific}.
\textit{Minimal pairs:} The neutral baseline images are generated from short scene descriptions written by Gemini 2.5 Flash.
We manually inspected all image pairs for whether the intended pole is shown while other aspects are preserved; the few failed edits (fewer than 5\%) were regenerated with a revised edit instruction or a stronger model.
\textit{Layers:} We steer early, middle and late ranges of four consecutive layers (eight for Emu3.5, which is twice as deep) at one quarter, one half and three quarters of the model's depth, e.g., layers 7--10, 16--19 and 25--28 of 36 in UniAR.
\textit{Evaluation prompts:} For each concept, we generate 20 neutral scene descriptions around one subject with Gemini 2.5 Flash (e.g., \textit{a wooden chair in a large empty room}). Image generation uses the description as is; for text generation, we iterate on various open-ended prompts, finding that \textit{``Describe your day in the city. Your description must include \{subject\}.''}
elicits diverse open-ended descriptions where steered concept can surface in whatever the model chooses to mention.
Full details are in \Cref{app:experimental-details}.

\section{Are steering directions shared?}
\label{sec:results-section}

\subsection{Measuring Cross-Modal Transfer}
\label{sec:measuring}
Given a concept $c$, a steering strength $\alpha$ and an evaluation prompt $p$, we write $\tilde{x}_{c,\alpha,p}$ for the steered output and $x_p$ for the unsteered output at $\alpha=0$.
We measure steering success with an LLM judge that compares the two, and further quantify cross-modal transfer by comparing how well a transferred vector steers with how well the intra-modality vector does.

\paragraph{Concept rate.} We use Gemini~3.7~Flash \citep{googledeepmind2026gemini37flash} as the judge. It sees $x_p$ and $\tilde{x}_{c,\alpha,p}$ in random order, without knowing which one is steered, and is asked which one shows more of a pole of the concept, for example \emph{happy} for emotion, with the asked pole chosen at random between the concept's two poles. The correct answer is the steered output when the judge is asked for the pole that $\alpha$ steers toward, and the baseline otherwise.
We score each pair $(x_p,\tilde{x}_{c,\alpha,p})$ as
\[
J\!\left(x_p,\tilde{x}_{c,\alpha,p}\right)=
\begin{cases}
1, & \text{if the judge picks the correct output},\\
0, & \text{if it picks the other output or answers \texttt{Tie}}.
\end{cases}
\]
The concept rate is the mean score over the $20$ prompts of a concept,
\[
\mathrm{CR}_{c,\alpha}
=
\frac{1}{|\mathcal{P}|}\sum_{p\in\mathcal{P}} J\!\left(x_p,\tilde{x}_{c,\alpha,p}\right),
\qquad |\mathcal{P}|=20 .
\]
To summarize a concept-rate curve across steering strengths $\alpha$ with one number, we take its normalized area under the curve, $\mathrm{AUC}_c\in[0,1]$.
An AUC of $1$ means the judge identified the concept in every output at every strength, and $0$ that it never did.~\footnote{AUC measures are commonly used as summary statistics in machine learning (e.g. for benchmark evaluation~\citep{subramani-etal-2025-simba}, classification (AUROC;~\citealp{marcum1960infotheory}), and calibration~\citep{JMLR:v11:el-yaniv10a, subramani-etal-2025-mice, subramani2026the}, and in the sciences~\citep{wagner1977bioavailability, hanley1982meaning}.}

\paragraph{Cross-modal recovery.} To judge \textit{transfer}, we measure how much of the intra-modality steering performance (i.e., $\mathrm{AUC}_c$) a transferred vector recovers.
We consider the intra-modality vector as a reference point since we expect a vector to steer the modality it was found in best.
Thus recovery is defined for image generation as $\mathrm{Rec}_{\text{img}}$ and for text generation as $\mathrm{Rec}_{\text{txt}}$:
\[
\mathrm{Rec}_{\text{img}}=\frac{\mathrm{AUC}_c(\text{txt}\to\text{img})}{\mathrm{AUC}_c(\text{img}\to\text{img})},
\qquad
\mathrm{Rec}_{\text{txt}}=\frac{\mathrm{AUC}_c(\text{img}\to\text{txt})}{\mathrm{AUC}_c(\text{txt}\to\text{txt})}.
\]
A recovery of $1$ means the transferred vector steers as well as the intra-modality one, $0$ means it does not steer at all, and above $1$ means it steers better.
Since each setting has its own strength range (\Cref{sec:applying-vectors}), recovery compares each vector over the strength ranges it tolerates, not at equal $\alpha$.
\paragraph{Further details.} The judge agrees with human raters at Cohen's $\kappa = 0.62$ \citep{cohen1960kappa}.
On top of concept rate, we separately report a metric for output quality, namely the fluency of steered text and the visual quality of steered images, adopted from \citep{wu2025axbench}.
Details in \Cref{app:experimental-details}.

\subsection{Results} 
\label{sec:results}

\begin{figure}[t]\centering
\includegraphics[width=\textwidth]{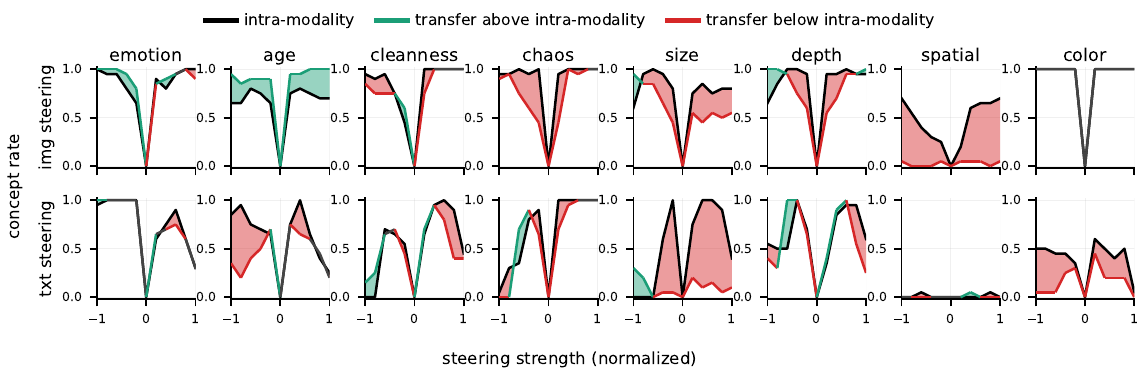}
\vspace{-3mm}
\caption{\textbf{Concept rate on UniAR \citep{peng2026unified} at middle layer range.}
Concept rate is the fraction of steered outputs in which an LLM judge finds the concept shifted toward the steered pole (\Cref{sec:measuring}), plotted against normalized steering strength.
Black curves show intra-modality steering (top: \qII, bottom: \qTT).
The colored areas indicate the gap between the two, either when transfer performs better (green) or worse (red). Recovery in \Cref{fig:recovery,tab:recovery} quantifies this comparison.
For most concepts, the transfer curve comes close to the intra-modality curve.
}
\label{fig:main-curves}
\end{figure}

\begin{figure}[t]
\centering
\begin{minipage}[c]{0.58\linewidth}
\centering
\includegraphics[width=\linewidth]{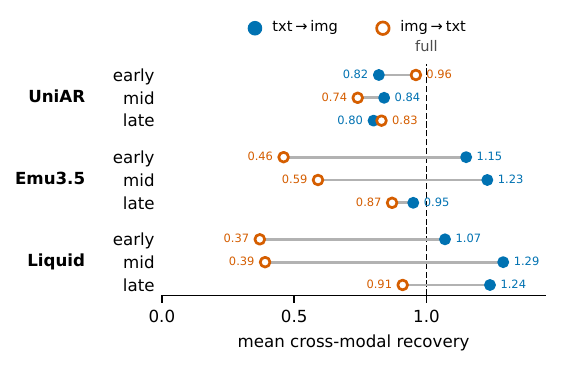}
\end{minipage}\hfill
\begin{minipage}[c]{0.38\linewidth}
\centering
\small
\setlength{\tabcolsep}{4pt}
\begin{tabular}{lcc}
\toprule
Concept & \qTI & \qIT \\
\midrule
emotion & 0.95 & 0.95 \\
age & 1.45 & 0.60 \\
cleanness & 0.97 & 0.69 \\
chaos & 1.06 & 0.70 \\
size & 1.12 & \textcolor{black!40}{0.32} \\
depth & 0.93 & 0.82 \\
spatial & \textcolor{black!40}{0.08} & \textcolor{black!40}{0.01} \\
color & 0.97 & 0.88 \\
\midrule
\rowcolor{black!7} mean & 1.04 & 0.68 \\
\bottomrule
\end{tabular}
\end{minipage}

\begin{minipage}[t]{0.58\linewidth}
\caption{\textbf{Cross-modal recovery} (averaged over concepts, showing steered models and layer ranges): ratio of transferred vector's AUC to intra-modality AUC of concept-rate curve (see \Cref{sec:measuring}). $1.0$ (dashed) means the transferred vector steers as well as the intra-modality one.}
\label{fig:recovery}
\end{minipage}\hfill
\begin{minipage}[t]{0.38\linewidth}
\captionof{table}{\textbf{Cross-modal recovery per concept}, averaged over models and layer ranges. Gray marks weak transfer (recovery $<0.5$).}
\label{tab:recovery}
\end{minipage}
\end{figure}

Across our three models and eight concepts, we report Cross-modal Recovery (both $\mathrm{Rec}_{\text{img}}$ and $\mathrm{Rec}_{\text{txt}}$) in \Cref{fig:recovery,tab:recovery}.
The concept-rate curves underlying these numbers are shown in \Cref{fig:main-curves} for UniAR and in the Appendix for Emu3.5 and Liquid (\Cref{fig:sweep-curves-play-emu35,fig:sweep-curves-play-liquid}).

\begin{figure}[t]\centering
\begin{minipage}[c]{0.56\textwidth}\centering
\includegraphics[width=\linewidth]{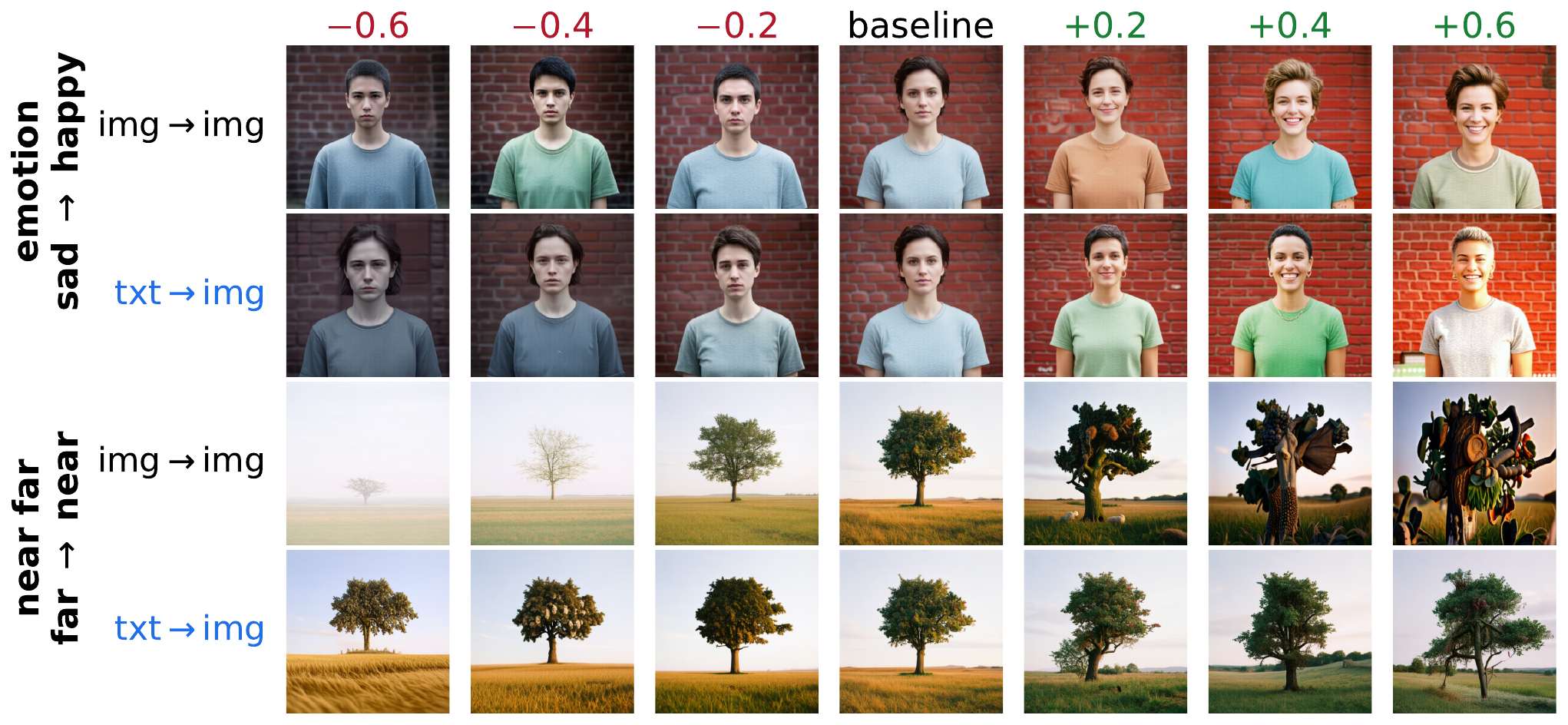}\par\vspace{1pt}
{\footnotesize\textbf{(a)} Image generation}
\end{minipage}\hfill
\begin{minipage}[c]{0.42\textwidth}\centering
\ktbox{\begin{tabularx}{\linewidth}{@{}l Y Y@{}}
 & \kp{sad}{happy} & \kp{clean}{dirty}\\
 \ka{black!55}{$0$} & \ktb{strolled through the city [...]} & \ktb{sleek wooden cabinets gleam [...]}\\[1pt]\midrule
\kms{pillNeg}{$-0.4$}{txt$\to$txt} & \kt{rain is falling [...] head down [...] lost in thought} & \kt{sleek white cabinets [...] minimalist design}\\\addlinespace[2.5pt]
\kms{pillPos}{$+0.4$}{txt$\to$txt} & \kt{sunny day [...] vibrant energy [...] perfect day} & \kt{messy kitchen, covered in food [...] covered in mud}\\\addlinespace[5pt]
\kms{pillNeg}{$-0.4$}{img$\to$txt} & \kt{serious expression [...] alone [...] empty streets} & \kt{sleek white cabinets [...] bright morning light}\\\addlinespace[2.5pt]
\kms{pillPos}{$+0.4$}{img$\to$txt} & \kt{sunny day [...] enjoying the vibrant atmosphere} & \kt{damp earth [...] scattered across the floor}\\
\end{tabularx}}\par\vspace{1pt}
{\footnotesize\textbf{(b)} Text generation}
\end{minipage}
\caption{\textbf{Cross-modal and intra-modal steering in UniAR} (middle layer range). We show examples for emotion, distance and cleanness steering.
In both modalities the transferred vector produces a similar shift as the intra-modality vector.}
\label{fig:qual}
\end{figure}

\paragraph{Steering vectors transfer in both directions.}
In all three models, vectors found in one modality also steer generation in the other for most concepts (\Cref{fig:recovery,tab:recovery}).
For every concept except spatial, there is a layer range at which text-derived vectors recover more than $80\%$ of the intra-modality effect, in all three models.
UniAR exhibits the strongest cross-modal steering effects: at middle layers, transferred vectors recover on average $84\%$ of the intra-modality effect on images and $74\%$ on text, and its concept-rate curves in \Cref{fig:main-curves} follow the intra-modality curves closely for most concepts.\footnote{We speculate this is due to the difference in image encoders (see \Cref{sec:experimental-details}).}
\qTI is strongest at middle layers in all three models; transfer into text peaks at early layers in UniAR and at late layers in Emu3.5 and Liquid, where their mean \qIT recovery reaches $0.87$ and $0.91$.
Examples for \qTI and \qIT are shown in \Cref{fig:qual}. 

\paragraph{\qTI is easier than \qIT.}
Text-derived vectors steer images for nearly every concept, and in Emu3.5 and Liquid even often better than intra-modality steering (\qII):
their \qTI recovery averages $1.10$ and $1.20$, i.e., above $1$.
We hypothesize this is mostly due to weaker image-derived vectors in these two models rather than particularly strong transfer:
\qII reaches an AUC of only $0.44$ and $0.43$ at middle layers, compared to $0.83$ in UniAR.
Practically, for these two models, text vectors are the better vectors for steering images, despite image-derived vectors being ``native'' to the space.
In the other direction, \qIT is only widely successful in UniAR, while more mixed in Emu3.5 and Liquid.
UniAR recovers on average up to $0.96$ of the intra-modality effect, whereas Emu3.5 and Liquid come close only at late layers  ($0.87$ and $0.91$).


\paragraph{Emotion transfers best, size and spatial worst.}
Emotion transfers most evenly, with a mean recovery of $0.95$ in both directions, and age transfers into images better than other concepts ($1.02$--$1.87$ across models and layer ranges).
Visual concepts tend to work better in the \qTI setting than in \qIT.
Size shows this most clearly with overall mean recovery of $1.12$ for \qTI but only $0.32$ for \qIT.
The only concept that fails across most settings is spatial (left-right), with moderate success only for \qII in UniAR (see curve in \Cref{fig:main-curves}) and minor effects for \qTT in Emu3.5 and Liquid.

\paragraph{Output quality.}
We check whether there is a strong trade-off between high concept rates and output quality.
With an AxBench-like judge~\citep{wu2025axbench}, we find that output quality noticeably drops only beyond the strengths where the concept rate plateaus (and, for text, then drops): at $0.4$ of the strength range, where the text concept rate peaks, text fluency is still at about three quarters of the level of unsteered outputs, and it collapses beyond that point (\Cref{app:axbench}).
On the image generation side the trade-off is smaller: image coherence is still at about $90\%$ of its unsteered level at $0.6$ of the range, where the concept rate has already reached over $90\%$ of its maximum.

\subsection{Qualitative analysis: modality-specific concepts}
\label{sec:qualitative}

So far we have studied 
concepts that 
occur in both language and vision, such as emotion or age, and some that are more central to vision, such as left-right.
We next study concepts that are 
unique to one modality, but 
harder to define 
in the other one. For example, what would an image look like 
if steered along a uniquely linguistic concept? 
We qualitatively inspect and discuss two examples for each modality in UniAR, as illustrated in \Cref{fig:teaser}.
Detailed sample outputs are in \Cref{app:native-examples}.

\paragraph{Language.}
We construct sets of minimal pairs in language that should elicit two purely linguistic steering axes:
Formal vs. colloquial language (``I need to go'' vs. ``I gotta go'') and language identity (``I love apples'' vs.``J'adore des pommes'').
We verify that these work as intended on language, e.g., producing more and more colloquial e-mails, or switching languages.
Despite our minimal pairs only defining the \textit{linguistic} aspect of formality and not, e.g., suits vs. sportswear, we find that this steering direction transfers to visual properties such as clothing, black-and-white vs. colorful, and luxurious minimalist style.
The English--French direction is also purely linguistic, translating the same semantic content into a different form.
Nonetheless, the steered images exhibit an increasingly French style that is historical and ornate rather than modern (mansard roofs, Art Nouveau posters, a vintage car), and, surprisingly, people express more feminine-presenting visual characteristics.
Writing in the image turns into French-looking but non-existent words.


\paragraph{Vision.}
It is less trivial to find exclusively visual concepts since any concept, almost by definition, has words associated to it or might not allow reasonable minimal pair construction.
However, we identify two successful case studies.
First, applying a steering vector found in images for the axis of low-to-high resolution makes generated text less (or more) detailed: fewer generated tokens, shorter sentences, mentions of ``blurry'' and even expressions of confusion\footnote{We confirm that overall shorter responses is not an artifact of steering: applying other steering vectors lead too much less reduction in length, if any at all.}; see \Cref{fig:teaser}.
We observe similar effects when applying a ``motion blur'' vector to text, with descriptions showing either calmer words on one end of the spectrum or faster movement-related ones on the end.
Finally, we successfully construct a steering vector for camera angle, i.e., whether the camera is positioned above the subject or near the floor; it steers images cleanly, but reaches text only in minor ways, shifting the next-token probability distribution without ever changing the predicted token.

\subsection{Geometry of steering vectors}
\label{sec:geometry}

Do text-derived and image-derived steering vectors point in similar directions?
And if that is the case, would their cosine similarity be predictive how well cross-modal steering works?

\begin{figure}[t]\centering
\includegraphics[width=\textwidth]{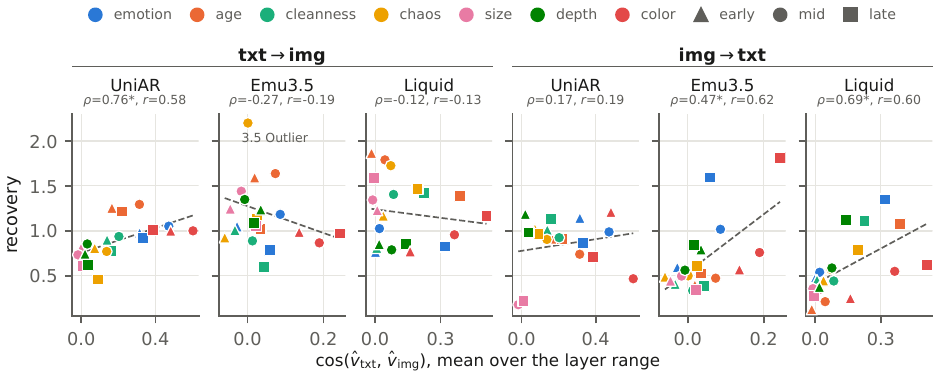} 
\vspace{-5mm}
\caption{\textbf{Cross-modal cosine similarity vs.\ cross-modal steering success.} Each point is one (concept, layer range): $x$ is cosine between text- and image-derived steering vectors; $y$ is our recovery metric (\Cref{sec:measuring}) for the given transfer direction (left: \qTI, right: \qIT). Spatial is excluded because it does not steer in most settings; $\rho$: Spearman, $r$: Pearson; $^*$ marks $p<0.05$.
}
\label{fig:recovery-vs-cos}
\end{figure}

\paragraph{Finding: Cosine similarities correlate with cross-modal steering (\Cref{fig:recovery-vs-cos}).}
For UniAR and Liquid, we find that cosine similarities for most concepts are significantly above zero\footnote{30 of the 42 (concept, layer range) points of UniAR and Liquid in \Cref{fig:recovery-vs-cos} lie more than two standard deviations ($0.03$) above random baseline.}, up to $0.60$ (UniAR) and $0.51$ (Liquid).
On the other hand, Emu3.5's steering vectors are close to zero cosine similarity for all concepts except color (maximum $0.24$).
Concretely, averaging across all (model, concept, layer range) points, we find a moderate correlation between cosine similarity and our main metric of steering transfer (\textit{recovery}, averaged over both transfer directions): Spearman $\rho=0.38$ ($p<0.01$), Pearson $r=0.22$ ($n=63$, the $21$ points per model of \Cref{fig:recovery-vs-cos}).
However, this correlation varies significantly across studied models and between \qIT vs. \qTI.
For example, in UniAR the correlation is strongest for \qTI transfer ($\rho=0.76$).
In all three models the correlating direction is the one whose \emph{intra-modality} steering is strongest (\qII in UniAR, \qTT in Liquid and Emu3.5), i.e.\ where the denominator of the recovery ratio is reliable.

\paragraph{Steering vectors show some alignment despite modality gap.}


For example, the happy--sad vectors found in text and in images point in somewhat similar directions, with a cosine similarity of $0.47$ in UniAR's mid layers.
Broadly, there could be two explanations for this:
1) The text and image activations for happy inputs are close, text and image activations for sad inputs are close, therefore the two difference vectors are close.
2) Text and image occupy separate subspaces but the two subspace share a similar structure, i.e., relations between the activations in each space are similar. \textbf{Setup:}
We test both explanations in UniAR's representation space: each concept pole in each modality (e.g., happy images) is summarized by the mean activation over its minimal-pair examples.
Activations are strongly anisotropic: two random activations of the same modality have a cosine similarity of about $0.75$ at the middle layers, and two from opposite modalities still $0.65$.
Therefore, we rescale them to unit length and subtract the overall mean before computing cosines (\Cref{app:geometry}); the steering vectors themselves are unaffected by this centering.

We find evidence for the second explanation:
Image and text occupy separate narrow spaces inside the model, while the relative directions between activations within each region exhibits some alignment.
If the first explanation were right, a happy image and a happy sentence would sit close together in the model.
However, in UniAR's middle layers, the average happy image is more than ten times closer to the average \emph{sad} image than to the average happy sentence, with similar or stronger trend for every concept.
On the other hand, the steering directions extracted in text and in images have a cosine similarity between $0.14$ and $0.60$ for five of the eight concepts, far above a random-direction baseline of $\pm0.03$, while size, near-far and spatial-lr are around zero.
Emu3.5 and Liquid show the same separation by modality.
Our findings illustrate that 
steering vectors can provide a separate avenue for quantifying whether a unified model is \textit{actually} unified.
Aligned directions despite separate subspaces is also related to modality-gap studies in contrastive embedding spaces \citep{liang2022mindthegap, zhang2024connectcollapsecorrupt} and the vector-offset regularities in word embeddings, i.e., the famous king-queen example \citep{mikolov2013linguisticregularities, park2024linearrepresentationhypothesisgeometry} and the Platonic Representation Hypothesis \citep{huh2024platonicrepresentationhypothesis, groger2026revisiting}.



\section{Related Work}
\label{sec:related-work-ref}


\paragraph{Unification and Alignment of Modalities.}
Approaches and tasks for connecting vision and language differ widely, from
multimodal language models \citep{lu2019vilbert,tsimpoukelli2021multimodal, alayrac2022flamingo}, text-conditioned image generation \citep{rombach2022ldm}, multimodal retrieval \citep{radford2021learning}, to the many variants of recent unified multimodal models \citep{chameleonteam2025chameleonmixedmodalearlyfusionfoundation,wang2024emu3nexttokenpredictionneed,deng2025emergingpropertiesunifiedmultimodal}.
Across all, it is widely studied how vision and language representations interact and align \citep{merullo2023linearly,neo2025interpretingvisual, wu2025semantichub}, ranging from modality gaps studies in CLIP models \citep{liang2022mindthegap} to circuit analyses and vocabulary space projections in multimodal LLMs \citep{nikankin2025same, krojer2026latentlens, feucht2026ocrheads}.
Only few works have studied internals of unified models, via attention patterns and embedding geometry \citet{cite_narrow_gate}, transfer of knowledge edits \citet{gao2026text} and steering vectors \citet{wang2026unifiedmultimodalmodelsthink}.
They often report a lack of unification \citep{cite_narrow_gate, gao2026text}, in contrast to our more positive findings.
Most similar to us is \citet{wang2026unifiedmultimodalmodelsthink}
who report cross-modal steering vectors in only one direction (\qTI).
They primarily study a different type of UMM, BAGEL \citep{deng2025emergingpropertiesunifiedmultimodal} with separate language and image generation weights and objectives,
while relying on minimal text inputs for both text and image vectors.
Finally, they study closed-form QA text outputs and not our setting of free-form text.

\paragraph{Visual steering.}
Most related to our transfer question, \citet{gan2026textualsteering} and \citet{deng2026causalprobinginternalvisual} each study one direction of transfer (text to image tokens and image to text, respectively) to alter the visual \emph{understanding} of MLLMs, which only take images as conditioning.
Other continuous attribute control in image generation \citep{baumann2025continuoussubjectspecific, ekin2026unreasonableeffectivenesstextembedding, dalva2024noiseclr, kamenetsky2025saedit, wang2026generalefficientsteering} likewise targets diffusion models rather than the shared space of a unified model.
Steering the visual-token stream of
an \emph{autoregressive} image generator, as we do, is rarely explored \citep{chrabaszcz2026conditionedactivationtransport, wang2026unifiedmultimodalmodelsthink}.

\paragraph{Textual steering.}
Identifying and applying linear and non-linear ``concept directions'' started with operating on embeddings themselves~\citep{mikolov2013distributed, bolukbasi2016man}.
Now, we operate in the latent spaces of language models for both exact and attribute-based steering~\citep{subramani2019can, subramani-etal-2022-extracting, you2026spherical}.
For these causal steering operations, data often comes from contrastive prompt pairs \citep{turner2024steeringlanguagemodelsactivation, rimsky2024contrastiveactivationaddition, zou2025representationengineeringtopdownapproach}, in-context task demonstrations~\citep{todd2024functionvectors}, or sparse-autoencoder features~\citep{huben2024sparseautoencoders}, all grounded in the linear representation hypothesis \citep{park2024linearrepresentationhypothesisgeometry}.

\section{Conclusion}
In this paper we have shown cross-modal steering vectors in both directions (\qTI and \qIT).
We also characterized and provided intuition how such cross-modal steering can manifest in the other modality.
As unified multimodal models become more capable and mature as a field, we see potential for methods that leverage this unified nature for model control.
If we identify a promising (steering) intervention to make models safer \citep{arditi2024refusal}, it should automatically transfer to visual capabilities; likewise, if we want to regularly update knowledge in the model, such knowledge edits should transfer to visual capabilities \citep{gao2026text}.
We encourage future work to expand our findings to other families of unified models and novel concepts.
Scientifically, we see unified models as a fruitful testbed for developing modality-agnostic interpretability tools and theories of multimodal interactions, as exemplified by our study of cross-modal steering.

\section*{Acknowledgements}
We would like to thank the TeamReddy lab at Mila for early brainstorming and feedback on the idea.
We are also grateful to Yossi Gandelsman for fruitful brainstorming sessions at earlier stages of the project.
DZ was supported by the McGill's SURA fellowship for part of the project.
BK is supported by the Vanier
Canada Graduate Scholarship.
TRS is supported by ARL, MIT-IBM Watson AI Lab, and Hyundai Motor Company, and specifically this research was supported by ISF research grant No. 4248/26.

\section*{Ethics Statement}
Our work studies internal steering directions of open-weight unified multimodal models on synthetic minimal pairs; we foresee no direct ethical concerns beyond
the plethora of general risks of controllable generation, including generation of harmful content targeting and/or exploiting vulnerable populations.

\section*{Reproducibility Statement}
\Cref{app:experimental-details} contains everything needed to reproduce our results:
the construction of the minimal pairs with all edit instructions,
the evaluation prompts, decoding settings per model, layer and strength selection,
the judge prompts and metrics, and the
human validation.
\Cref{app:geometry} details the geometry analyses.
Code is at \url{https://github.com/saintsauceee/cross-modal-steering}, and the minimal pairs,
steering vectors, generations and judge scores at \url{https://huggingface.co/saintsauce}.

\section*{AI Use Statement}
We used generative AI tools throughout this project, always in a back-and-forth with the authors, and not simply adopting what they suggest.

\textit{Research and experiments.} We used coding agents to implement the methods and to run most experiments reported in the paper, including cleaning and reformatting the generated data. Some methodological decisions, the hypotheses we tested, and how prior work has approached the problem were also discussed with AI tools. We never blindly adopted their suggestions, it was a long back-and-forth in which we often disagreed with the AI assistant and often stuck to our own scientific intuition. AI assistants also helped interpret results and describe qualitative patterns in steered outputs; every such interpretation was checked against the underlying numbers and generations by the authors. We did not review agent-written code line by line. Instead, we discussed the code and its outputs with the agent, re-ran experiments, and sanity-checked results (e.g., for strange or implausible values) before adopting them into the paper.

\textit{Data and evaluation.} We used generative AI models to automatically generate the minimal pairs for concept directions, including the French sentences of the language-identity pairs, and the evaluation prompts (\Cref{app:minimal-pairs,app:eval-prompts}). The authors inspected the generated pairs and measured how minimal they are before using them. An LLM (Gemini 3.7 Flash) served as the judge for all reported metrics; its agreement with human raters is reported in \Cref{app:human}.

\textit{Writing, figures and literature.} The final paper is the full responsibility of, and was primarily driven by, us authors. We sometimes adopted isolated sentence-level suggestions from AI assistants, especially for sentences reporting technical details or for rephrasing a suboptimal phrase, and used them much more sparingly for writing than for coding and experiments. AI tools helped create the figures, and helped identify and summarize related work; when an AI tool suggested citations, we checked them against its original source.

We did not use AI for developing theoretical models or for formulating or proving mathematical claims (not applicable here). All AI-assisted content was reviewed by the authors, who take responsibility for the final content of this work.


\bibliography{references,iclr2027_conference}
\bibliographystyle{iclr2027_conference}

\clearpage
\appendix
\crefalias{section}{appendix}
\crefalias{subsection}{appendix}
\begin{center}{\LARGE\bfseries Appendix}\end{center}
\vspace{0.2em}
\addtocontents{toc}{\protect\setcounter{tocdepth}{1}}
{\makeatletter\@starttoc{toc}\makeatother}
\vspace{1em}

\section{Experimental Details}
\label{app:experimental-details}

\paragraph{Concepts.}
Our concepts are chosen from those commonly studied in either the language or vision community. Emotion (sad/happy) is commonly steered in language models \citep{turner2024steeringlanguagemodelsactivation, zou2025representationengineeringtopdownapproach}, and age (young/old) and expression are canonical targets of latent-direction editing in image generation \citep{shen2020interfacegan, gandikota2024conceptsliders, baumann2025continuoussubjectspecific}. Cleanness and chaos are semantic concepts that describe the state of a whole scene. In text a single word expresses them, while in an image they are spread over many objects, which makes them an interesting test of whether one direction can capture both. The visual concepts describe basic properties of objects in a scene, such as their size, depth, spatial position or color. We expect spatial (left/right) to be the hardest concept to steer, since spatial relations are a well-documented failure mode of image generation models and VLMs \citep{chatterjee2024gettingitright, kamath-etal-2023-whats}.

\subsection{Minimal Pairs}
\label{app:minimal-pairs}

Each steering vector is the mean difference in hidden states over a concept's minimal pairs, which are either all text or all images. We describe how both kinds of pair are constructed below.

\paragraph{Text pairs.} We use Gemini 2.5 Flash to generate $N=100$ distinct subjects that fit the concept and fill each into the template ``a \slot{\{subject\}} is \slot{\{pole\}}'', once with each pole. For people in the color concept, the template is ``a \slot{\{subject\}} is dressed in \slot{\{pole\}}''. Each text sample is thus an unpunctuated short declarative sentence ending in the pole itself, so the two sentences of a text minimal pair differ only there, for instance \textit{a chef is happy} and \textit{a chef is sad}. \Cref{tab:text-pairs} gives more examples.

\begin{table}[H]
\centering
\small
\caption{Example text minimal pairs. Each subject gives one pair that differs only in the pole word.}
\label{tab:text-pairs}
\vspace{10pt}
\begin{tabular}{@{}lll l@{}}
\toprule
Concept & Example subjects & $-$ pole & $+$ pole \\
\midrule
emotion & teacher, chef, coach & \textit{a teacher is \textcolor{negcol}{sad}} & \textit{a teacher is \textcolor{poscol}{happy}} \\
age & engineer, cook, merchant & \textit{an engineer is \textcolor{negcol}{young}} & \textit{an engineer is \textcolor{poscol}{old}} \\
cleanness & oven, fork, glove & \textit{an oven is \textcolor{negcol}{clean}} & \textit{an oven is \textcolor{poscol}{dirty}} \\
chaos & street, hospital, party & \textit{a street is \textcolor{negcol}{peaceful}} & \textit{a street is \textcolor{poscol}{chaotic}} \\
size & flower, pen, statue & \textit{a flower is \textcolor{negcol}{small}} & \textit{a flower is \textcolor{poscol}{large}} \\
depth & flower, pen, statue & \textit{a flower is \textcolor{negcol}{far away}} & \textit{a flower is \textcolor{poscol}{nearby}} \\
spatial & flower, pen, statue & \textit{a flower is \textcolor{negcol}{on the left}} & \textit{a flower is \textcolor{poscol}{on the right}} \\
color & hoodie, bus, chair & \textit{a hoodie is \textcolor{negcol}{blue}} & \textit{a hoodie is \textcolor{poscol}{red}} \\
\bottomrule
\end{tabular}
\end{table}

\begin{figure}[H]
\centering
\includegraphics[width=\textwidth]{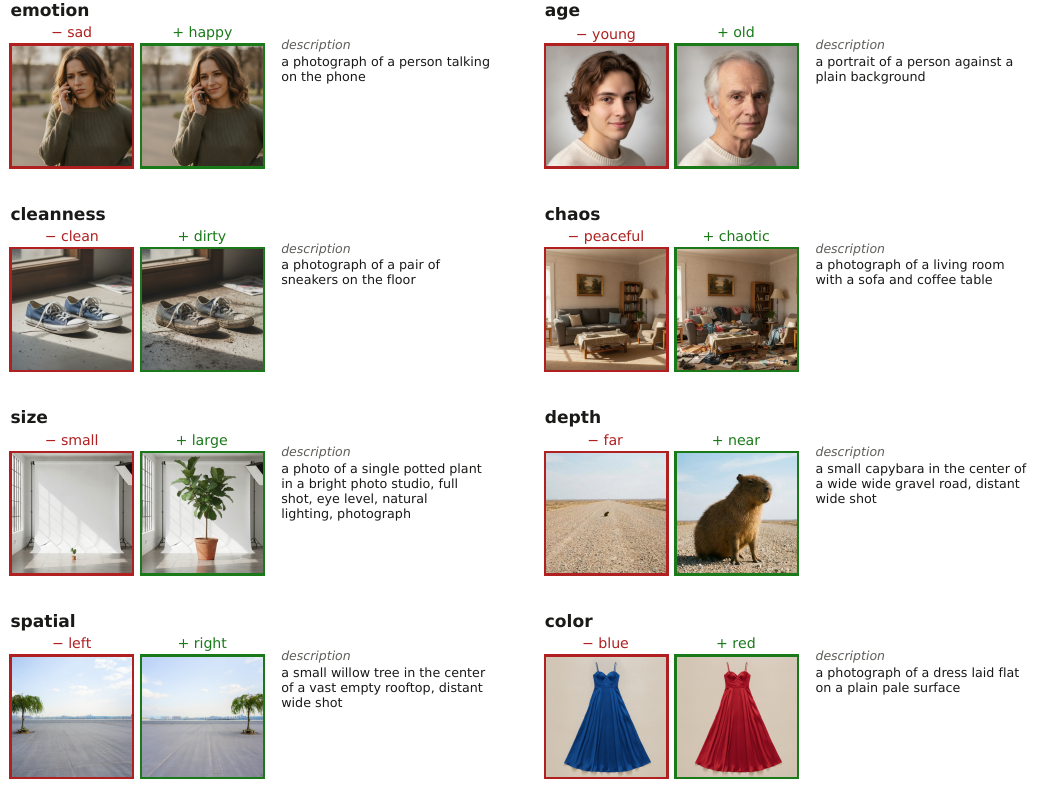}
\caption{One image pair per concept, with the description used to generate its baseline image.}
\label{fig:minimal-pairs}
\end{figure}

\newpage
\paragraph{Image pairs.} 
An image pair consists of two images of the same scene that differ only in the pole. 
We start from 100 short image descriptions per concept, written by Gemini 2.5 Flash. 
Each names a subject and, if needed, what it is doing and where. 
We generate a baseline image from each description and edit it toward each pole. For most concepts, both steps use Gemini 2.5 Flash Image. 
However, depth, size and spatial edits are harder and Gemini 2.5 Flash Image often gets them wrong, so for these we use Gemini 3 Pro Image and Reve. 
\Cref{fig:minimal-pairs} shows one image pair per concept with its description. The edit instructions are listed below for each concept and pole.

\begingroup\tcbset{before skip=10pt, after skip=10pt}
\begin{jpromptbox}[Edit instruction: emotion]
\ttfamily\footnotesize\raggedright\setlength{\baselineskip}{12pt}
\textbf{$-$ (sad)}: Subtly modify this image to make the person look sad. Everything else, such as identity, pose, hair, clothing, background and lighting, should be preserved exactly as before. Please only change the facial expression.\\[9pt]
\textbf{$+$ (happy)}: Subtly modify this image to make the person look happy. Everything else, such as identity, pose, hair, clothing, background and lighting, should be preserved exactly as before. Please only change the facial expression.
\end{jpromptbox}

\begin{jpromptbox}[Edit instruction: age]
\ttfamily\footnotesize\raggedright\setlength{\baselineskip}{12pt}
\textbf{$-$ (young)}: Modify this image to make the person look much younger, a young child or young teenager, with smooth wrinkle-free skin, a fuller rounder face and bright clear eyes. Everything else, such as the person, pose, framing, clothing and background, should be preserved exactly as before. Please only change the apparent age.\\[9pt]
\textbf{$+$ (old)}: Modify this image to make the person look much older, elderly, in their 70s or 80s, with deep wrinkles, grey or white thinning hair, sagging skin and age spots. Everything else, such as the person, pose, framing, clothing and background, should be preserved exactly as before. Please only change the apparent age.
\end{jpromptbox}

\begin{jpromptbox}[Edit instruction: cleanness]
\ttfamily\footnotesize\raggedright\setlength{\baselineskip}{12pt}
\textbf{$-$ (clean)}: Make everything in this image look spotless, freshly cleaned, polished and pristine, with no dirt, dust or stains anywhere. Everything else, such as the objects, layout, framing, camera angle and composition, should be preserved exactly as before. Please only change how clean or dirty the surfaces look.\\[9pt]
\textbf{$+$ (dirty)}: Make everything in this image look dirty, grimy, dusty and stained, covered in mud, dirt, grease and grime. Everything else, such as the objects, layout, framing, camera angle and composition, should be preserved exactly as before. Please only change how clean or dirty the surfaces look.
\end{jpromptbox}

\begin{jpromptbox}[Edit instruction: chaos]
\ttfamily\footnotesize\raggedright\setlength{\baselineskip}{12pt}
\textbf{$-$ (peaceful)}: Modify this scene to look peaceful, calm and serene: orderly and tidy, open and uncluttered, still and quiet. Everything else, such as the scene, layout, viewpoint and subjects, should be preserved exactly as before. Please only change how tidy or chaotic the scene is.\\[9pt]
\textbf{$+$ (chaotic)}: Modify this scene to look chaotic, turbulent and disordered: cluttered and messy, objects scattered, knocked over and in disarray, crowded and frantic. Everything else, such as the scene, layout, viewpoint and subjects, should be preserved exactly as before. Please only change how tidy or chaotic the scene is.
\end{jpromptbox}

\begin{jpromptbox}[Edit instruction: size]
\ttfamily\footnotesize\raggedright\setlength{\baselineskip}{12pt}
\textbf{$-$ (small)}: Make the \slot{\{subject\}} much smaller, tiny, only a few centimetres, taking up just a small part of the scene. Keep exactly one \slot{\{subject\}}, do not add or duplicate it. Everything else, such as the spot, the room, the camera and the size and position of everything around it, should be preserved exactly as before. Please only change the size of the \slot{\{subject\}}.\\[9pt]
\textbf{$+$ (large)}: Make the \slot{\{subject\}} a giant \slot{\{subject\}}, several times larger than normal, towering and reaching high up near the top of the frame. Keep exactly one \slot{\{subject\}}, do not add or duplicate it. Everything else, such as the spot, the room, the camera and the size and position of everything around it, should be preserved exactly as before. Please only change the size of the \slot{\{subject\}}.
\end{jpromptbox}

\begin{jpromptbox}[Edit instruction: depth]
\ttfamily\footnotesize\raggedright\setlength{\baselineskip}{12pt}
\textbf{$-$ (far)}: Make the single main subject much smaller and more distant, so it occupies only a tiny part of the frame. Keep exactly one subject visible, do not add, remove, clone or duplicate it. Everything else, such as the subject, background scene, camera framing, lighting and style, should be preserved exactly as before. Please only change how close the subject is.\\[9pt]
\textbf{$+$ (near)}: Make the single main subject much larger so it fills most of the frame, like a close-up. Keep exactly one subject visible, do not add, remove, clone or duplicate it. Everything else, such as the subject, background scene, camera framing, lighting and style, should be preserved exactly as before. Please only change how close the subject is.
\end{jpromptbox}

\begin{jpromptbox}[Edit instruction: spatial]
\ttfamily\footnotesize\raggedright\setlength{\baselineskip}{12pt}
\textbf{$-$ (left)}: Move the single main subject to the far left side of the frame. Keep exactly one subject visible, do not add, remove, clone or duplicate it. Everything else, such as the subject, background scene, camera framing, lighting and style, should be preserved exactly as before. Please only change where the subject is in the frame, left or right.\\[9pt]
\textbf{$+$ (right)}: Move the single main subject to the far right side of the frame. Keep exactly one subject visible, do not add, remove, clone or duplicate it. Everything else, such as the subject, background scene, camera framing, lighting and style, should be preserved exactly as before. Please only change where the subject is in the frame, left or right.
\end{jpromptbox}

\begin{jpromptbox}[Edit instruction: color]
\ttfamily\footnotesize\raggedright\setlength{\baselineskip}{12pt}
\textbf{$-$ (blue)}: Modify this image so the subject is clearly blue, its main surfaces repainted, re-dyed or remade in a saturated blue. Everything else, such as the subject's shape, material, placement and scale, the background, framing, camera and lighting, should be preserved exactly as before. The rest of the scene keeps its original colours and brightness. Please only change the colour of the subject.\\[9pt]
\textbf{$+$ (red)}: Modify this image so the subject is clearly red, its main surfaces repainted, re-dyed or remade in a saturated red. Everything else, such as the subject's shape, material, placement and scale, the background, framing, camera and lighting, should be preserved exactly as before. The rest of the scene keeps its original colours and brightness. Please only change the colour of the subject.
\end{jpromptbox}
\endgroup

\subsection{Evaluation Prompts}
\label{app:eval-prompts}
Every concept gets its own set of $20$ evaluation prompts, with subjects and scenes that make sense for it. We reuse them across all layer configurations and steering strengths.

Each evaluation prompt is a description of one to three complete sentences that names a subject and describes the scene around it. For image generation, we use the full description (\slot{\{image description\}}) to give the model more to condition on and to keep more of the scene fixed. For text generation, we keep only its main subject (\slot{\{subject\}}) and place it in an open instruction, leaving the model free to describe the scene in as much visual detail as an image would show (\Cref{app:generation}). \Cref{tab:eval-prompts} shows a few examples. Subjects are not limited to people, but include other subjects where the concept can show, such as toys and figurines with a face for emotion, or a bike for age. 

\begin{table}[H]
\centering
\footnotesize
\caption{Example evaluation prompts. Image generation uses the full description, and text generation uses only its subject, shown in bold.}
\label{tab:eval-prompts}
\vspace{10pt}
\begin{tabular}{@{}p{2.2cm}p{8.4cm}p{2.9cm}@{}}
\toprule
Concept & Image description & Text subject \\
\midrule
emotion & A head-and-shoulders portrait of \textbf{a person with short cropped hair}, wearing a plain crew-neck shirt and holding a relaxed, neutral expression. They stand in front of a weathered red brick wall in soft natural daylight. It is shot as a sharp, realistic photograph with an even, unremarkable mood. & \textit{a person with short cropped hair} \\\addlinespace[3pt]
cleanness & A photograph of \textbf{a modern kitchen with wooden cabinets}, stone countertops, and a stainless steel sink. Even natural daylight from a window falls across the counters, backsplash, and floor. It is a plain, realistic interior photograph with no people. & \textit{a modern kitchen with wooden cabinets} \\\addlinespace[3pt]
size & A single \textbf{traffic cone} in the center of a wide empty asphalt parking lot with faded painted markings, under flat even daylight. & \textit{a traffic cone} \\\addlinespace[3pt]
color & A photograph of \textbf{a car parked} alone on a wide empty asphalt lot, the car centred and seen from the side, under soft even daylight and a pale overcast sky. A plain, realistic photograph with no people. & \textit{a parked car} \\\addlinespace[3pt]
\bottomrule
\end{tabular}
\end{table}

\newpage
We generate the evaluation prompts with Gemini 2.5 Flash, using the prompt below. It asks for scenes that are neutral with respect to the concept, so that the unsteered outputs are not biased toward either pole.

\begin{jpromptbox}[Evaluation prompt generation]
\ttfamily\footnotesize\raggedright\setlength{\baselineskip}{12pt}
Write 20 short descriptions of photographs. Each description is one to three sentences about one main subject, saying what it is doing and where, when. Be sure to keep all subjects distinct.\\[6pt]
The descriptions will be used to test the concept \slot{\{concept\}}, whose two poles are \slot{\{positive pole\}} and \slot{\{negative pole\}}. Every description must be neutral on this concept: do not describe the subject or the scene as either pole, and avoid any word that leans toward one of them.\\[6pt]
Respond as JSON only, giving the subject phrase of each description as its own field:\\
\{\\
~~"concept": "\slot{\{concept\}}",\\
~~"prompts": [\\
~~~~\{\\
~~~~~~"subject": "<subject>",\\
~~~~~~"description": "<one to three sentences>"\\
~~~~\},\\
~~~~...\\
~~]\\
\}
\end{jpromptbox}

\subsection{Generation and Decoding}
\label{app:generation}
\paragraph{Text generation.} All judged text outputs are generated with the same instruction, shown below in each model's own chat format. We iterated on this instruction, since description-style prompts made the model focus on a single aspect of the scene, or refuse or output JSON, often leaving little room for steering. The final instruction lets the model freely mention many concepts and visual associations, so that a steered concept can surface in whatever the model chooses to mention, while the shared subject keeps the two modalities comparable. Only the subject phrase, taken from \Cref{app:eval-prompts}, changes from one prompt to the next. The instruction states that no image is provided, since a unified model may otherwise expect one. We steer the subject phrase and every generated token.
\begin{cpromptbox}[Text prompt: UniAR]
\ttfamily\footnotesize\raggedright\setlength{\baselineskip}{12pt}
\imtag{im\_start}\textbf{system}\\
\textbf{You are a helpful assistant.}\imtag{im\_end}\\[4pt]
\imtag{im\_start}\textbf{user}\\
No image is provided; write from imagination. Answer in two or three plain sentences, no title, no list. Describe your day in the city. Your description must include \steer{\slot{\{subject\}}}.\imtag{im\_end}\\[4pt]
\imtag{im\_start}\textbf{assistant}\\
\steer{[generated tokens]}\promptlegend
\end{cpromptbox}

\begin{cpromptbox}[Text prompt: Emu3.5]
\ttfamily\footnotesize\raggedright\setlength{\baselineskip}{12pt}
\imtag{extra\_203}\textbf{You are a helpful assistant.}\\
\textbf{USER:} No image is provided; write from imagination. Answer in two or three plain sentences, no title, no list. Describe your day in the city. Your description must include \steer{\slot{\{subject\}}}.\\[4pt]
\textbf{ASSISTANT:} \imtag{extra\_100}\steer{[generated tokens]}\promptlegend
\end{cpromptbox}

\begin{cpromptbox}[Text prompt: Liquid]
\ttfamily\footnotesize\raggedright\setlength{\baselineskip}{12pt}
\vtag{start\_of\_turn}\textbf{user}\\
No image is provided; write from imagination. Answer in two or three plain sentences, no title, no list. Describe your day in the city. Your description must include \steer{\slot{\{subject\}}}.\vtag{end\_of\_turn}\\[4pt]
\vtag{start\_of\_turn}\textbf{model}\\
\steer{[generated tokens]}\promptlegend
\end{cpromptbox}

\paragraph{Image generation.} Each full image description (\Cref{app:eval-prompts}) is given to the model in its own image-generation format, shown below. The steered tokens are only the generated visual tokens, not the prompt.

\begin{cpromptbox}[Image prompt: UniAR]
\ttfamily\footnotesize\raggedright\setlength{\baselineskip}{12pt}
\imtag{im\_start}\textbf{user}\\
\slot{\{image description\}}\vtag{image\_gen}\textbf{ generate image 16 16}\imtag{im\_end}\\[4pt]
\imtag{im\_start}\textbf{assistant}\\
\imtag{vision\_start}\steer{[generated visual tokens]}\promptlegend
\end{cpromptbox}

\begin{cpromptbox}[Image prompt: Emu3.5]
\ttfamily\footnotesize\raggedright\setlength{\baselineskip}{12pt}
\imtag{extra\_203}\textbf{You are a helpful assistant for t2i task.}\\
\textbf{USER:} \slot{\{image description\}}\\[4pt]
\textbf{ASSISTANT:} \imtag{extra\_100}\steer{[generated visual tokens]}\promptlegend
\end{cpromptbox}

\begin{cpromptbox}[Image prompt: Liquid]
\ttfamily\footnotesize\raggedright\setlength{\baselineskip}{12pt}
\slot{\{image description\}} \textbf{Generate an image based on this description.}\\[4pt]
\vtag{boi}\steer{[generated visual tokens]}\promptlegend
\end{cpromptbox}

The decoding settings of every sweep are listed in \Cref{tab:sweeps}.

\begin{table}[H]
\centering
\footnotesize
\setlength{\tabcolsep}{5pt}
\caption{The strength grid is set per model and setting (\Cref{app:layer-strength-selection}). We keep each model's default decoding settings other than the number of decoder steps and the image resolution for practical reasons, and fix the seed at $42$.}
\label{tab:sweeps}
\vspace{10pt}
\begin{tabular}{@{}lllp{6.4cm}@{}}
\toprule
Model & Setting & $\alpha$ grid & Decoding \\
\midrule
UniAR  & \qTT, \qIT & $\pm100$, step $20$ & greedy, $96$ tokens \\
\addlinespace[3pt]
       & \begin{tabular}[t]{@{}l@{}}\qII\\\qTI\end{tabular} & \begin{tabular}[t]{@{}l@{}}$\pm10$, step $2$\\$\pm15$, step $3$\end{tabular} & $16\times16$ tokens, guidance $1.5$, temperature $1.0$, SD3 decoder with $16$ steps and guidance $1.5$, $544\times544$ image \\
\midrule
Emu3.5 & \qTT, \qIT & $\pm25$, step $5$ & greedy, $96$ tokens \\
\addlinespace[3pt]
       & \begin{tabular}[t]{@{}l@{}}\qII\\\qTI\end{tabular} & \begin{tabular}[t]{@{}l@{}}$\pm1.5$, step $0.3$\\$\pm5$, step $1$\end{tabular} & $32\times32$ tokens, guidance $2.0$, top-$k$ $10{,}240$, top-$p$ $1.0$, temperature $1.0$, VQ decoder, $512\times512$ image \\
\midrule
Liquid & \qTT, \qIT & $\pm25$, step $5$ & greedy, $96$ tokens \\
\addlinespace[3pt]
       & \begin{tabular}[t]{@{}l@{}}\qII\\\qTI\end{tabular} & \begin{tabular}[t]{@{}l@{}}$\pm0.5$, step $0.1$\\$\pm1.5$, step $0.3$\end{tabular} & $32\times32$ tokens, guidance $7.0$, top-$k$ $4{,}096$, top-$p$ $0.96$, temperature $0.99$, VQ decoder, $512\times512$ image \\
\bottomrule
\end{tabular}
\end{table}

\subsection{Layer and Steering Strength}
\label{app:layer-strength-selection}
\paragraph{Layer configurations.}
Traditionally, steering operates on a single layer, at one or all timesteps depending on the task~\citep{subramani-etal-2022-extracting}, but more recent work has found success steering multiple or even all layers~\citep{zou2025representationengineeringtopdownapproach, gaintseva2025casteer, sun2025layernavigator}.
Different layers encode different linguistic properties and later layers tend to encode more complex features in neural network models~\citep{ethayarajh-2019-contextual, tenney-etal-2019-bert, rogers-etal-2020-primer, skean2025layer, li-subramani-2026-model}.
We therefore compare early, middle and late layers without choosing one over the others.\footnote{In interpretability, splitting layers into early, middle, and late is common practice~\citep{subramani2026the}.}
We steer seven layer configurations per model: three single layers, three contiguous layer ranges and all layers, listed in \Cref{tab:layer-configs}.
Each layer range spans four consecutive layers around its matching single layer, or eight in Emu3.5 since it has about twice as many layers ($64$, against $36$ and $28$), so that a layer range covers a similar share of the depth in every model.
A transferred vector is injected at the same layers it was extracted from.

\begin{table}[H]
\centering
\small
\caption{Layer configurations for each model ($0$-indexed). Single layers are evenly spaced in depth, each contiguous layer range spans the layers around its matching single layer, and the seventh configuration steers all layers.}
\label{tab:layer-configs}
\vspace{10pt}
\begin{tabular}{lcccccccc}
\toprule
 & & \multicolumn{3}{c}{Single layer} & & \multicolumn{3}{c}{Contiguous layer range} \\
\cmidrule(lr){3-5}\cmidrule(lr){7-9}
Model & Depth & early & middle & late & & early & middle & late \\
\midrule
UniAR  & 36 & 9  & 18 & 27 & & 7--10  & 16--19 & 25--28 \\
Emu3.5 & 64 & 16 & 32 & 48 & & 12--19 & 28--35 & 44--51 \\
Liquid & 28 & 7  & 14 & 21 & & 5--8   & 12--15 & 19--22 \\
\bottomrule
\end{tabular}
\end{table}

\paragraph{Steering strength $\alpha$.}
Each model and setting has its own grid of steering strengths $\alpha$, with five evenly spaced values on each side of the unsteered $\alpha = 0$ (\Cref{tab:sweeps}). To find the range, we first sweep a fine grid of $\alpha$ on a few evaluation prompts at the middle layer range. We then pick the range so that, across all concepts, the one or two largest $|\alpha|$ on each side produce degenerate samples, such as incoherent text or broken images. This grid is used for all seven layer configurations.

\subsection{Judge Setup and Prompts}
\label{app:concept-rate}
\paragraph{Judge.} The judge is Gemini~3.7~Flash with a low thinking level and one sample per comparison, and a JSON response schema restricts its verdict to \texttt{A}, \texttt{B} or \texttt{Tie}. It never sees any steering hyperparameter, such as $\alpha$, the layer configuration or the steering vector. We validate it against human reviewers in \Cref{app:human}.

\paragraph{Prompt.} The prompts are shown below, where \slot{\{pole\}} is the asked pole, \slot{\{anchor\}} the opposite pole and \slot{\{answer\}} the rater's answer. We prompt the judge few-shot with ten example pairs for each modality. They are held-out pairs on which the reviewers agreed, and their answers are spread evenly across the steered output, the baseline and a tie. The question names one of the two poles at random, so the judge cannot score well just by recognizing steering artifacts.

\begin{jpromptbox}[Concept rate judge: image pair]
\ttfamily\fontsize{9}{12}\selectfont\raggedright
You are a careful annotator comparing two model outputs.\\[10pt]
Here are examples of how this rater judges such pairs. The image shows numbered rows; in each row the left panel is A and the right panel is B:\\[2.5pt]
Row 1: Which represents more of \{pole\} (or less of \{anchor\})? Answer: \{answer\}\\[2.5pt]
\ldots\\[2.5pt]
Row 10: Which represents more of \{pole\} (or less of \{anchor\})? Answer: \{answer\}\\[2.5pt]
[example grid image, rows 1A\textbar 1B to 10A\textbar 10B]\\[10pt]
Now judge this new pair the same way:\\[2.5pt]
A: [image A]\\[2.5pt]
B: [image B]\\[10pt]
Which represents more of \{pole\} (or less of \{anchor\})?\\[2.5pt]
Choose A or B only if one clearly and obviously shows more of the attribute at a glance. If the difference is small, faint, or you have to look hard, answer Tie.\\[2.5pt]
Judge only the attribute - ignore style, grammar, and quality. A candidate that looks odd, broken, or repetitive can still clearly show the attribute (e.g. it repeats the trait), and is not automatically a Tie; decide only by whether the attribute itself is clearly stronger in one.\\[2.5pt]
Answer with one word (A, B, or Tie).
\end{jpromptbox}

\begin{jpromptbox}[Concept rate judge: text pair]
\ttfamily\fontsize{9}{12}\selectfont\raggedright
You are a careful annotator comparing two model outputs.\\[10pt]
Here are examples of how this rater judges such pairs:\\[2.5pt]
[10 examples, in the format below]\\[10pt]
Now judge this pair the same way:\\[10pt]
Candidate A:\\[2.5pt]
[text A]\\[10pt]
Candidate B:\\[2.5pt]
[text B]\\[10pt]
Which represents more of \{pole\} (or less of \{anchor\})?\\[2.5pt]
Choose A or B only if one clearly and obviously shows more of the attribute at a glance. If the difference is small, faint, or you have to look hard, answer Tie.\\[2.5pt]
Judge only the attribute - ignore style, grammar, and quality. A candidate that looks odd, broken, or repetitive can still clearly show the attribute (e.g. it repeats the trait), and is not automatically a Tie; decide only by whether the attribute itself is clearly stronger in one.\\[2.5pt]
Answer with one word (A, B, or Tie).
\end{jpromptbox}

\begin{jpromptbox}[Text example format]
\ttfamily\fontsize{9}{12}\selectfont\raggedright
Example:\\[2.5pt]
Candidate A:\\[2.5pt]
[example text A]\\[10pt]
Candidate B:\\[2.5pt]
[example text B]\\[10pt]
Which text represents more of \{pole\} (or less of \{anchor\})?\\[2.5pt]
Choose A or B, or Tie if it's not obvious.\\[2.5pt]
Answer with one word (A, B, or Tie).\\[2.5pt]
Answer: \{answer\}
\end{jpromptbox}

\subsection{DreamSim Pole Score (Images)}
\label{app:dreamsim}
As a complement to the concept rate metric for images, we compare each generated image with two references, a positive-pole and a negative-pole edit of the same scene. With $d$ the cosine distance in DreamSim, a learned perceptual image similarity \citep{fu2023dreamsim}, the pole score of an image $I$ with references $R^{+}, R^{-}$ is
\[
\mathrm{pole}(I) \;=\; 100 \cdot \frac{\; d\!\left(I,\, R^{-}\right) \;}{\; d\!\left(I,\, R^{+}\right) \,+\, d\!\left(I,\, R^{-}\right) \;},
\]
so $100$ means $I$ matches the positive-pole reference, $0$ the negative-pole reference, and $50$ is equidistant. The references for a prompt are the model's own unsteered image, minimally edited toward each pole with Gemini 3 Pro Image, following the same method as the minimal pairs.

\begin{figure}[H]\centering
\includegraphics[width=\textwidth]{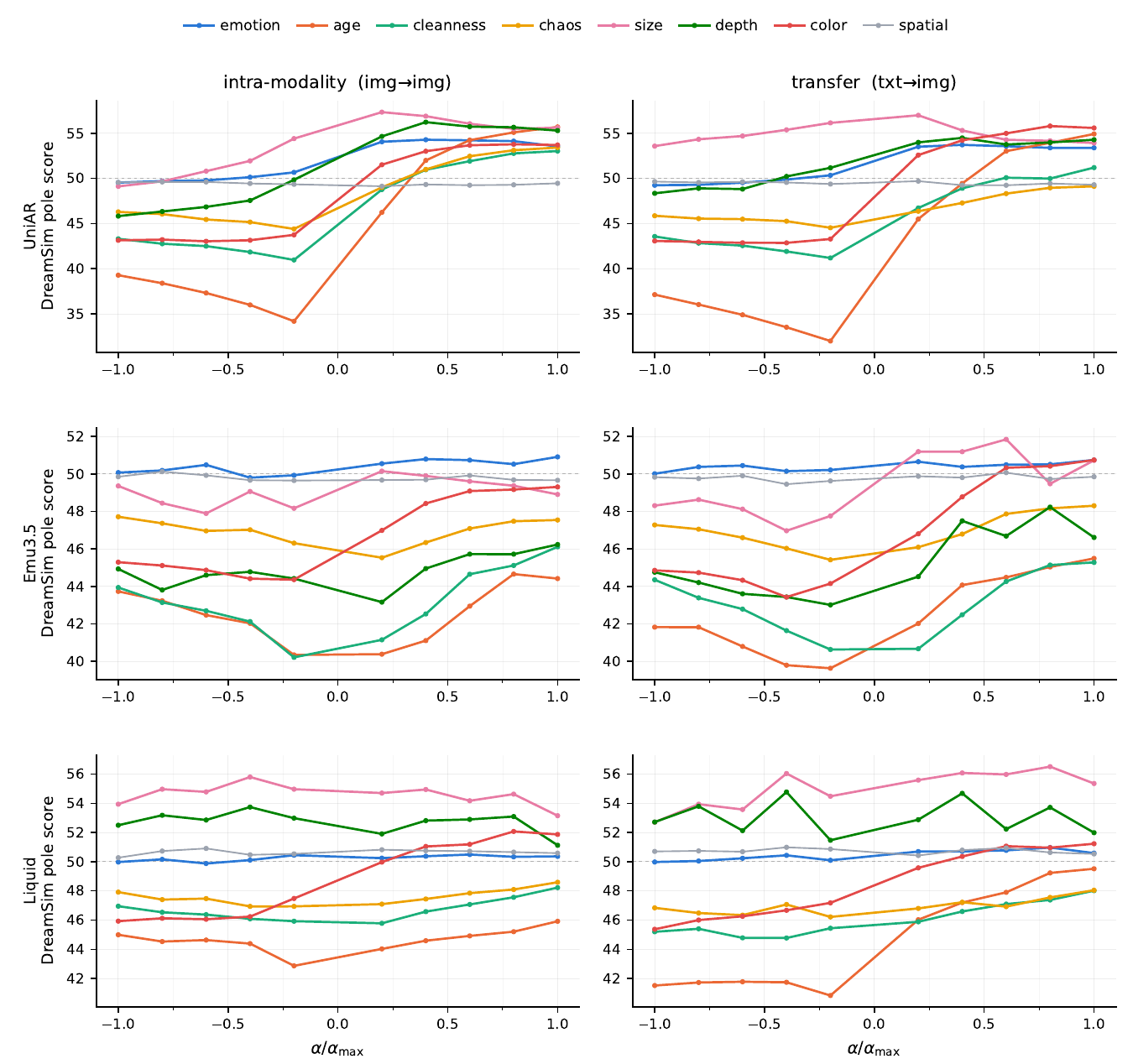}
\caption{DreamSim pole score for image-derived (left) and text-derived (right) vectors.}
\label{fig:dreamsim}
\end{figure}

\Cref{fig:dreamsim} plots the pole score against normalized strength for the two image-output settings. It rises with $\alpha$ for the concepts that steer and stays flat at $50$ for spatial. The same rise appears when a text-derived vector steers image generation.

UniAR moves the furthest, with age strongest and every concept except spatial rising, apart from size under the text-derived vector. In Emu3.5 and Liquid the shifts are only a few points: color moves most with both vectors, and age moves mainly with the text-derived vector. In Liquid, size and depth sit above $50$ at every strength without moving, so their offset is not steering.

\subsection{Fluency and Coherence Judges}
\label{app:axbench}
\textbf{Text.} We use an AxBench-style pointwise fluency judge \citep{wu2025axbench}. Gemini~3.7~Flash scores each output $0$, $1$ or $2$ for fluency and sees only the output, never the prompt, the strength or the layer. We score all $22{,}400$ steered text outputs per model plus one unsteered baseline per concept and prompt. The prompt is shown below.

\begin{promptbox}[Text fluency]
You are evaluating the linguistic fluency of a language model output.

Model output:
{text}

Rate the fluency and coherence of the output.

Use the following scale:

0 = The output is substantially incoherent, ungrammatical, or otherwise difficult to understand.
1 = The output is understandable but contains noticeable issues with grammar, coherence, or naturalness.
2 = The output is fluent, coherent, grammatical, and natural.

Focus only on linguistic quality. Do not judge whether the output follows the original instruction or expresses the target concept.

Return JSON only:
{"score": 0 | 1 | 2, "reasoning": "<one concise sentence>"}
\end{promptbox}

\textbf{Images.} For images we build the counterpart of the fluency judge: the same model sees only the image, with no prompt, concept or strength, and returns one of three verdicts, \emph{intact}, \emph{partly broken} or \emph{broken}, which we map to $2$/$1$/$0$. We use verdict words rather than a numeric scale, which returned $1$ for nearly every image. We score all $3{,}200$ steered images per model at the middle layer range and the $160$ unsteered baselines. The prompt is shown below.

\begin{promptbox}[Image coherence]
You are checking an image generated by a model for visible breakage.

Answer with one of three verdicts:

"intact": the image is coherent and well-formed. This is the expected verdict for a typical generated image; the usual look of AI-generated images (smoothness, softness, a synthetic feel) is not breakage.
"partly_broken": the content is recognizable, but there are clear artifacts, distortions, or incoherent elements, such as malformed hands or faces, merged or duplicated objects, garbled text, or corrupted regions.
"broken": severe artifacts or garbled textures dominate, or the content cannot be recognized.

Only visible breakage matters. Do not judge whether the image matches any description or expresses any concept, and do not judge style, aesthetics, or realism.

Return JSON only:
{"reasoning": "<one concise sentence naming any breakage you see, or stating that there is none>", "verdict": "intact" | "partly_broken" | "broken"}
\end{promptbox}

\begin{figure}[H]\centering
\includegraphics[width=\textwidth]{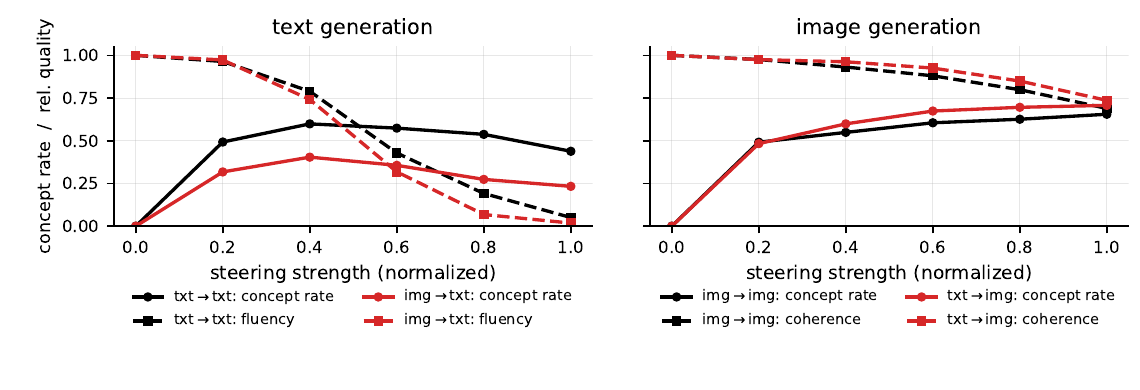}
\caption{Concept rate (solid) and output quality (dashed) against normalized strength at the middle layer range, averaged over models, concepts, prompts and both signs of $\alpha$. Quality is text fluency (left) and image coherence (right), relative to the unsteered outputs.}
\label{fig:quality}
\end{figure}

\textbf{Results.} \Cref{fig:quality} plots both quality scores next to the concept rate against normalized strength, each relative to the model's unsteered outputs. Text fluency holds up to about $0.4$ of the range, where the concept rate peaks, and collapses beyond it. Image coherence declines gently over the whole range while the concept rate keeps rising. Transferred and intra-modality vectors degrade quality similarly, except that image-derived vectors cost text fluency more than text-derived ones in Emu3.5 and Liquid at middle strengths.

\subsection{Human Validation}
\label{app:human}
\begin{figure}[H]\centering
\includegraphics[width=0.85\textwidth]{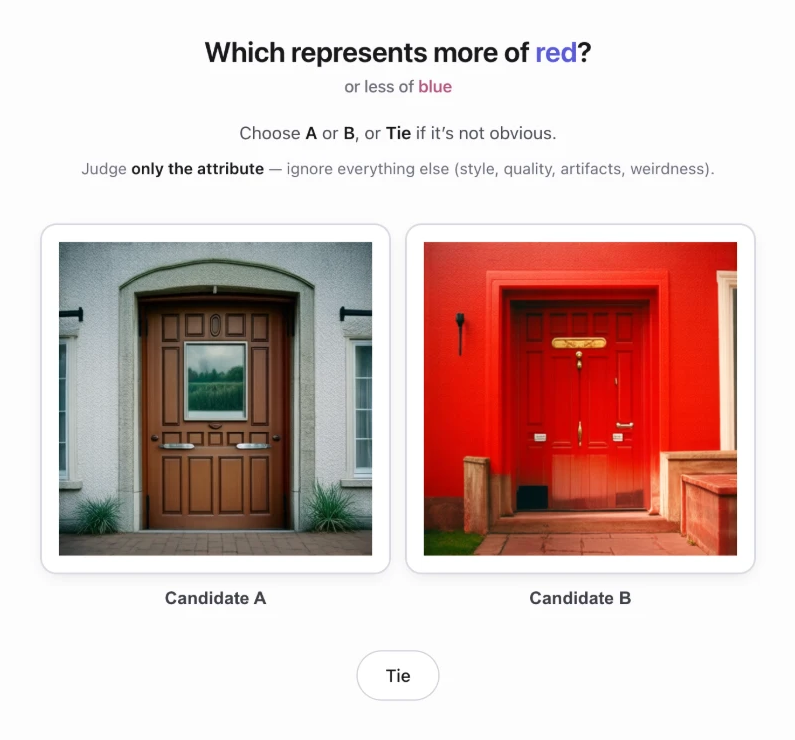}
\caption{The annotation interface shown to human reviewers. It names the asked pole and its opposite. Two model outputs, the unsteered baseline and the steered output, are shown as Candidate A and B in random order. The human reviewer answers A, B or Tie.}
\label{fig:annotation-ui}
\end{figure}

To check that the automated concept-rate judge tracks human judgment, we collect human annotations on $320$ held-out pairs, $10$ per concept and setting across eight concepts, none of which is part of the evaluated sweeps. Each pair receives three independent reviews. Reviewers face the same forced-choice task as the judge: the steered output and its $\alpha=0$ baseline shown as A and B in \emph{randomized order}, with the asked pole also \emph{randomized} ($\approx$50\% the steered pole, $\approx$50\% the opposite), and are blind to $\alpha$, the layer, and which output is steered. \Cref{fig:annotation-ui} shows the annotation interface.

Since the A/B order and the asked pole are random, we recode each answer as agreeing with the steering direction, going against it, or a tie, and combine the three reviews of a pair by majority vote, breaking three-way splits at random. Cohen's $\kappa$ between the judge and this majority label is $0.62$ over the $320$ pairs.

\section{Detailed results}
\label{app:full-sweep}
\begin{figure}[H]
\centering
\includegraphics[width=\textwidth]{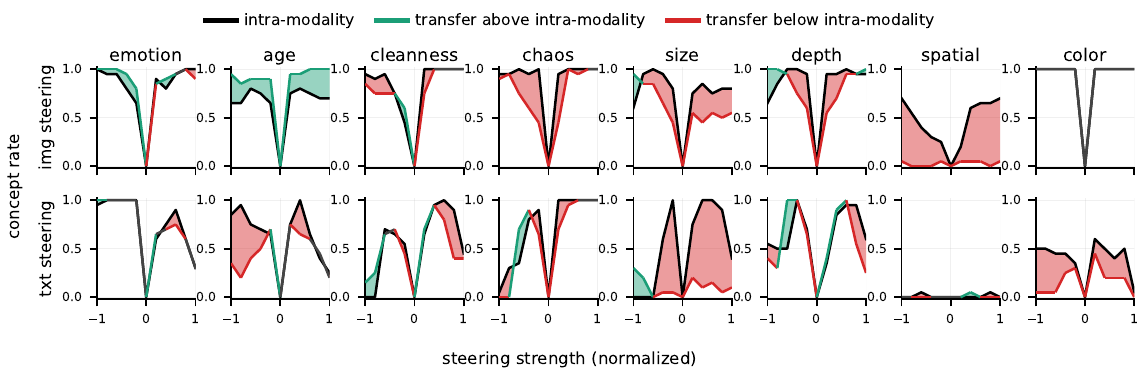}
\caption{UniAR at the middle layer range, same as \Cref{fig:main-curves}, repeated for comparison. The black curve is the intra-modality vector and the colored curve the transferred one. The shaded area between them is their difference in AUC, green where the transferred vector steers more and red where it steers less.}
\label{fig:sweep-curves-play-uniar}
\end{figure}

\begin{figure}[H]
\centering
\includegraphics[width=\textwidth]{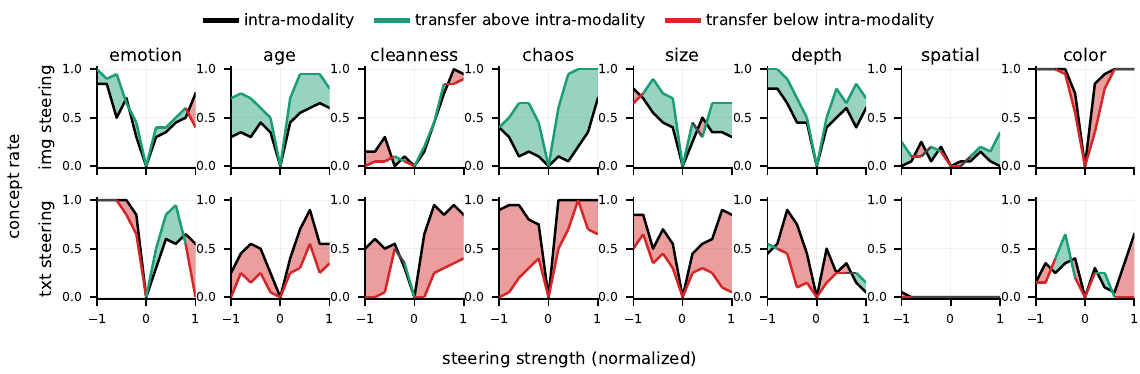}
\caption{Emu3.5 at the middle layer range, same layout as \Cref{fig:main-curves}.}
\label{fig:sweep-curves-play-emu35}
\end{figure}

\begin{figure}[H]
\centering
\includegraphics[width=\textwidth]{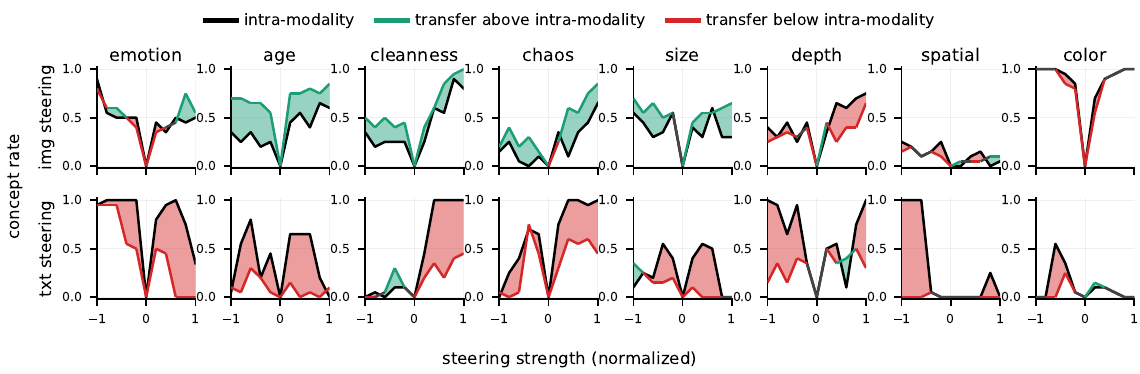}
\caption{Liquid at the middle layer range, same layout as \Cref{fig:main-curves}.}
\label{fig:sweep-curves-play-liquid}
\end{figure}

\begin{figure}[H]
\centering
\includegraphics[width=\textwidth]{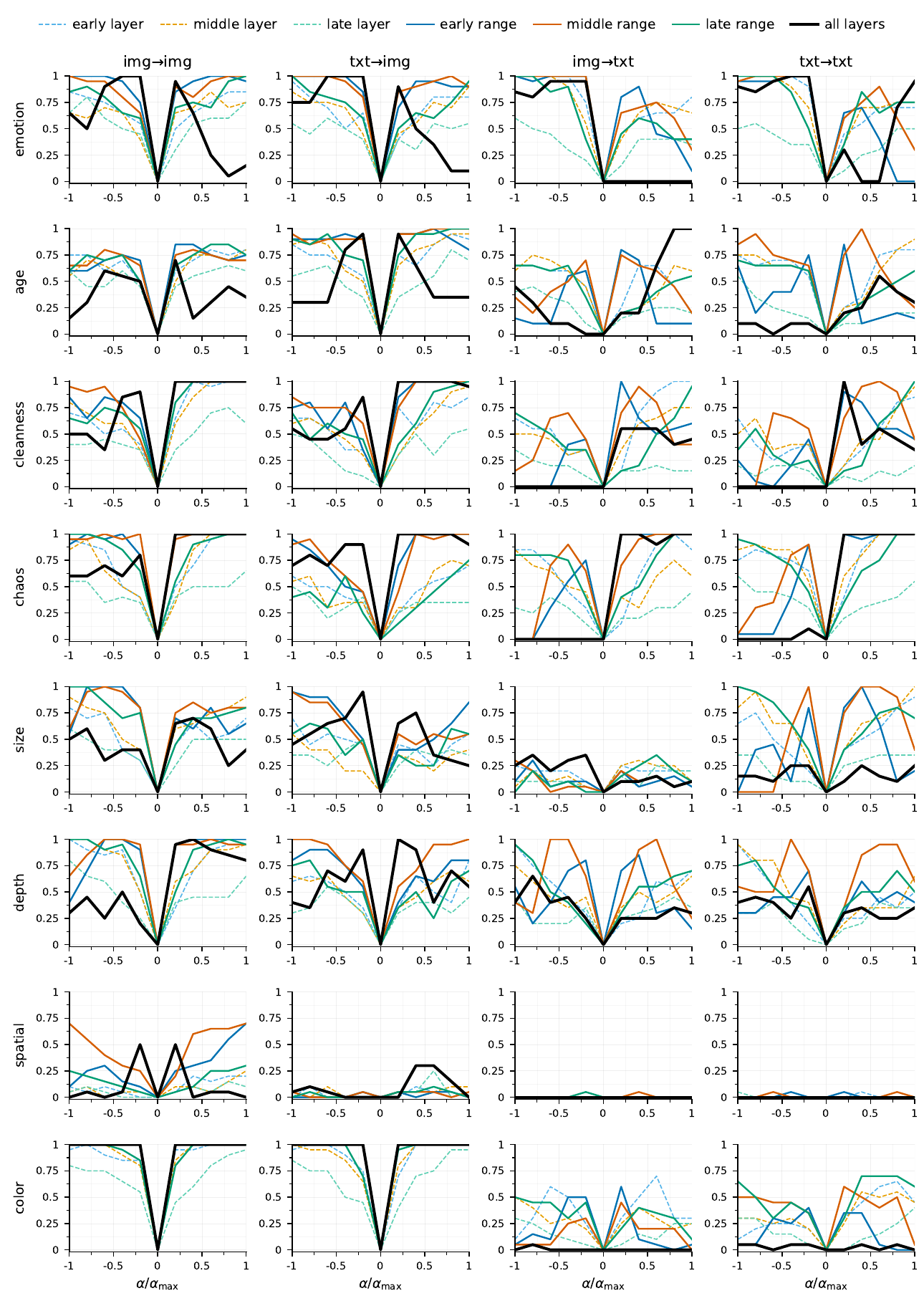}
\caption{UniAR, every concept (rows) and setting (columns), one line per layer configuration. Single layers are light and dashed, layer ranges solid, one hue per depth (blue early, orange middle, green late), and all layers is the thick black line. Every curve starts at rate $0$ at $\alpha = 0$.}
\label{fig:sweep-curves}
\end{figure}

\begin{figure}[H]
\centering
\includegraphics[width=\textwidth]{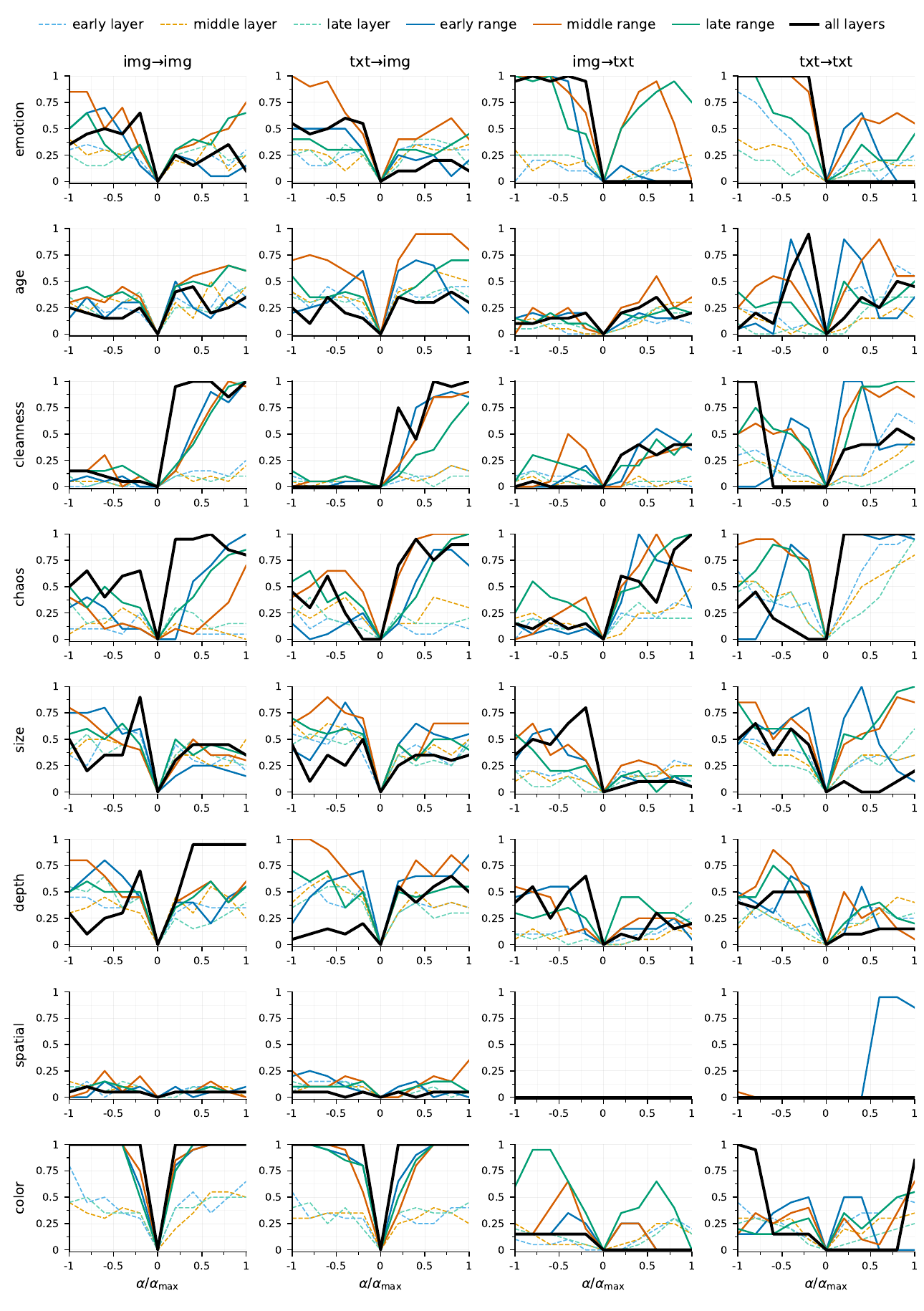}
\caption{Same as \Cref{fig:sweep-curves} for Emu3.5.}
\label{fig:sweep-curves-emu35}
\end{figure}

\begin{figure}[H]
\centering
\includegraphics[width=\textwidth]{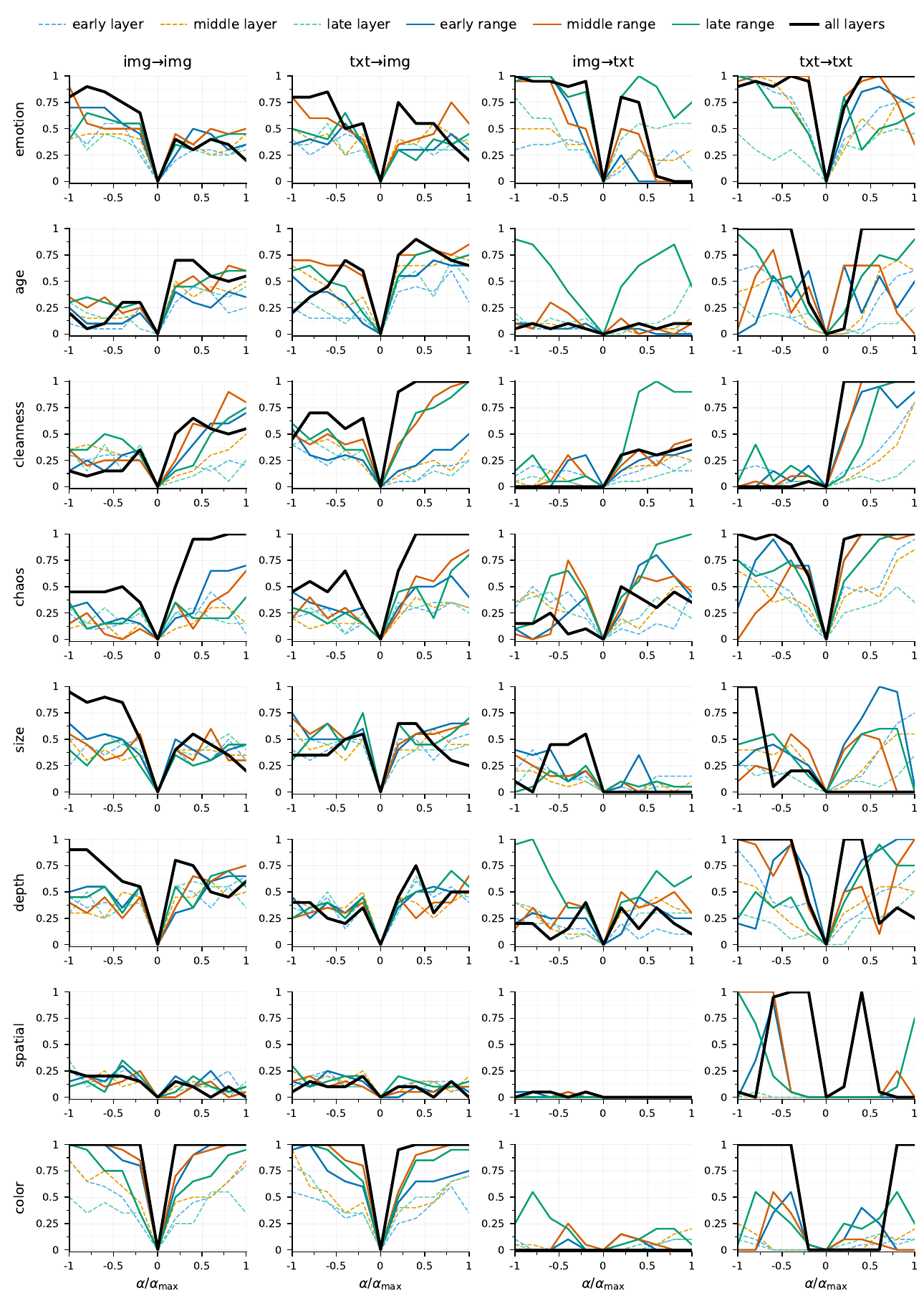}
\caption{Same as \Cref{fig:sweep-curves} for Liquid.}
\label{fig:sweep-curves-liquid}
\end{figure}

This section shows the concept-rate curves of every concept and setting for all three models. \Cref{fig:sweep-curves-play-uniar,fig:sweep-curves-play-emu35,fig:sweep-curves-play-liquid} compare intra-modality and transferred vectors at the middle layer range, in the layout of \Cref{fig:main-curves}, with UniAR repeated for comparison. \Cref{fig:sweep-curves,fig:sweep-curves-emu35,fig:sweep-curves-liquid} show all seven layer configurations. Additionally, \Cref{tab:recovery-full} gives the cross-modal recovery of every concept, model and layer range.

\begin{table}[H]
\centering
\small
\setlength{\tabcolsep}{4pt}
\caption{\textbf{Cross-modal recovery}:
How much of intra-modality steering is \textit{recovered} by the transferred vector, defined as ratio of transfer AUC to the intra-modality AUC of concept-rate curve (see \Cref{fig:main-curves,sec:measuring}): $1.0$ means the transferred vector steers as well as intra-modality, $0$ no transfer; values above $1.0$ mean transferred vector steers better than intra-modality.
Cells are blank where intra-modality steering itself barely works (intra-modality AUC $<0.2$), so the ratio is not meaningful.
Gray marks weak transfer (recovery $<0.5$); bold marks the mean of the best layer range, per model and setting.}
\label{tab:recovery-full}
\vspace{10pt}
\begin{tabular}{lllcccccccc|>{\columncolor{black!7}}c}
\toprule
Model & Layers & Direction & emotion & age & cleanness & chaos & size & depth & spatial & color & mean \\
\midrule
\multirow{6}{*}{UniAR} & early & \qTI & 0.98 & 1.26 & 0.90 & 0.81 & 0.81 & 0.74 & \textcolor{black!40}{0.09} & 1.00 & 0.82 \\
 &  & \qIT & 1.14 & 0.91 & 1.01 & 1.02 & \textcolor{black!40}{0.25} & 1.19 & -- & 1.21 & \textbf{0.96} \\
\cmidrule(lr){2-12}
 & middle & \qTI & 1.05 & 1.29 & 0.94 & 0.77 & 0.73 & 0.85 & \textcolor{black!40}{0.07} & 1.00 & \textbf{0.84} \\
 &  & \qIT & 0.99 & 0.74 & 0.92 & 0.90 & \textcolor{black!40}{0.18} & 0.96 & -- & \textcolor{black!40}{0.47} & 0.74 \\
\cmidrule(lr){2-12}
 & late & \qTI & 0.92 & 1.22 & 0.78 & \textcolor{black!40}{0.46} & 0.61 & 0.62 & -- & 1.01 & 0.80 \\
 &  & \qIT & 0.87 & 0.91 & 1.13 & 0.97 & \textcolor{black!40}{0.22} & 0.98 & -- & 0.71 & 0.83 \\
\midrule
\multirow{6}{*}{Emu3.5} & early & \qTI & 1.05 & 1.59 & 1.01 & 0.93 & 1.25 & 1.24 & -- & 0.99 & 1.15 \\
 &  & \qIT & 0.59 & \textcolor{black!40}{0.40} & \textcolor{black!40}{0.41} & \textcolor{black!40}{0.49} & \textcolor{black!40}{0.44} & 0.79 & \textcolor{black!40}{0.00} & 0.57 & 0.46 \\
\cmidrule(lr){2-12}
 & middle & \qTI & 1.18 & 1.64 & 0.89 & -- & 1.44 & 1.35 & -- & 0.87 & \textbf{1.23} \\
 &  & \qIT & 1.02 & \textcolor{black!40}{0.47} & \textcolor{black!40}{0.33} & 0.50 & 0.50 & 0.56 & -- & 0.76 & 0.59 \\
\cmidrule(lr){2-12}
 & late & \qTI & 0.79 & 1.02 & 0.60 & 1.13 & 1.06 & 1.09 & -- & 0.97 & 0.95 \\
 &  & \qIT & 1.59 & 0.52 & \textcolor{black!40}{0.38} & 0.61 & \textcolor{black!40}{0.34} & 0.84 & -- & 1.81 & \textbf{0.87} \\
\midrule
\multirow{6}{*}{Liquid} & early & \qTI & 0.76 & 1.87 & 0.81 & 1.17 & 1.23 & 0.85 & -- & 0.77 & 1.07 \\
 &  & \qIT & \textcolor{black!40}{0.48} & \textcolor{black!40}{0.13} & \textcolor{black!40}{0.45} & \textcolor{black!40}{0.45} & \textcolor{black!40}{0.35} & \textcolor{black!40}{0.38} & -- & -- & 0.37 \\
\cmidrule(lr){2-12}
 & middle & \qTI & 1.03 & 1.79 & 1.40 & 1.73 & 1.34 & 0.79 & -- & 0.96 & \textbf{1.29} \\
 &  & \qIT & 0.54 & \textcolor{black!40}{0.21} & \textcolor{black!40}{0.44} & 0.57 & \textcolor{black!40}{0.36} & 0.59 & \textcolor{black!40}{0.02} & -- & 0.39 \\
\cmidrule(lr){2-12}
 & late & \qTI & 0.83 & 1.39 & 1.42 & 1.47 & 1.59 & 0.85 & -- & 1.17 & 1.24 \\
 &  & \qIT & 1.35 & 1.08 & 1.11 & 0.78 & \textcolor{black!40}{0.27} & 1.12 & -- & 0.62 & \textbf{0.91} \\
\midrule
\rowcolor{black!7}
 & all & \qTI & 0.95 & 1.45 & 0.97 & 1.06 & 1.12 & 0.93 & \textcolor{black!40}{0.08} & 0.97 & 1.04 \\
\rowcolor{black!7}
\multirow{-2}{*}{mean} &  & \qIT & 0.95 & 0.60 & 0.69 & 0.70 & \textcolor{black!40}{0.32} & 0.82 & \textcolor{black!40}{0.01} & 0.88 & 0.68 \\
\bottomrule
\end{tabular}
\end{table}

Every sweep follows the same recipe, described in \Cref{app:experimental-details}. For each model, we generate one output for every combination of concept, setting, layer configuration, strength and prompt, and judge it against its unsteered baseline with the concept-rate judge.

\section{Details for: geometry of steering vectors}
\label{app:geometry}

\begin{figure}[H]\centering
\includegraphics[width=\textwidth]{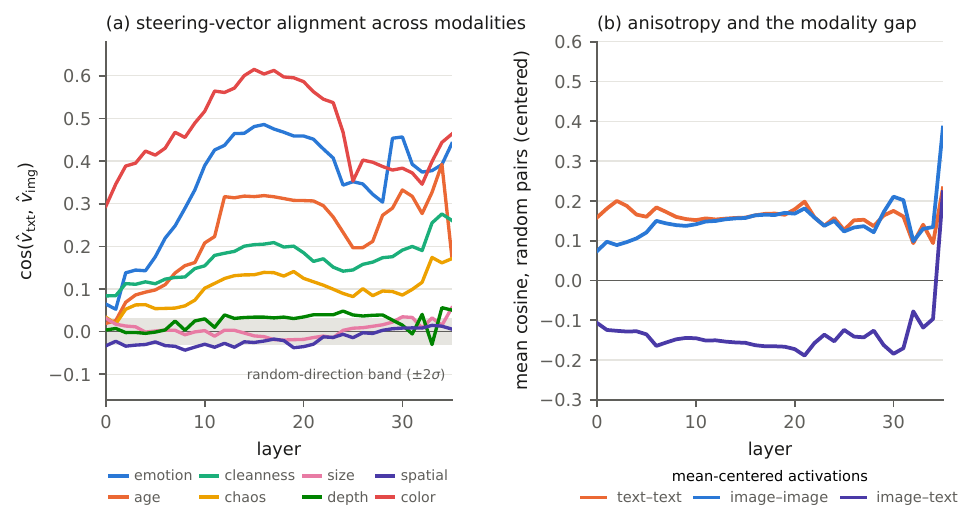}
\caption{\textbf{(a)} Per-layer cosine similarity between each concept's textual and visual steering vectors in UniAR; the gray band is $\pm2\sigma$ for random directions in $d{=}4096$ ($\sigma{=}0.016$). \textbf{(b)} Mean cosine similarity between random pairs of single-token activations across layers of UniAR, within text, within image, and across modalities, after centering (midpoint of the two modality means subtracted). The cross-modal cosine is negative after centering: the modality gap of \Cref{sec:geometry}.}
\label{fig:cosine}
\end{figure}

Here we provide details behind \Cref{sec:geometry} with full per-model results. \Cref{tab:geometry-correlations} shows the correlation between cosine similarity and steering success for every model and transfer direction (either with recovery score or AUC). \Cref{tab:anisotropy-models} gives the anisotropy numbers for all three models, and \Cref{tab:objects-models} the per-concept comparison of ``same pole'' (happy sentences, happy images) versus ``same direction'' (image steering vector for happy, text steering vector for happy). \Cref{fig:cosine} shows the per-layer cosine between the text and image steering vectors of each concept.

\paragraph{Cosine between steering vectors.}
We compare the same per-layer unit vectors that are injected during steering. A layer range is summarized by the mean of its per-layer cosines. As a baseline, the cosine between two random directions is zero on average, with a standard deviation of $1/\sqrt{d}$, which is $0.016$ for UniAR, $0.014$ for Emu3.5 and $0.018$ for Liquid. We call a cosine above the baseline if it exceeds two standard deviations, $2/\sqrt{d}$.

\paragraph{Activations and centering.}
For the modality-gap analysis, each sentence and each image is represented by one vector per layer, pooled the same way the steering vectors are built (last token for sentences, mean over image tokens for images), and a concept pole by the mean over its $100$ examples. Before comparing across modalities we subtract the midpoint between the image mean and the text mean, so that neither modality's sample size affects the center.  We rescale every activation to unit length before centering, because text activations are longer than image activations in Emu3.5 and Liquid, about $19\times$ and $3\times$ per token, and would otherwise dominate the comparison. In UniAR the two have nearly the same length, so this step changes little. The anisotropy numbers use single-token activations rather than pooled ones. All steering vectors are stable across random halves of their minimal pairs (split-half cosine at least $0.91$), except the emotion and spatial image vectors of Emu3.5 and Liquid ($0.45$ to $0.84$).
UniAR's activations at various layers exhibit a strong narrow cone effect (\textit{anisotropy}) \citep{ethayarajh-2019-contextual, liang2022mindthegap}: two random text or image activations at the middle layers (16--19 of 36) have an average cosine similarity of $0.75$ and $0.74$ respectively, while even two random activations from opposite modalities still have $0.65$ (\Cref{tab:anisotropy-models}). Rescaling to unit length and mean-centering is a common method to counteract anisotropy \citep{mu2018allbutthetop, cai2021isotropy}. Note that such centering leaves the steering vectors themselves unchanged: each is a difference of two means within one modality, $\hat v = \operatorname{mean}(h^+) - \operatorname{mean}(h^-)$, so any constant offset cancels.

\begin{table}[H]
\centering
\small
\caption{Correlation between the cosine similarity of the text and image steering vectors and cross-modal steering, per model and transfer direction. Each cell uses $21$ points ($7$ concepts $\times$ $3$ layer ranges; spatial excluded because it does not steer in most settings). Left: against recovery, as in \Cref{fig:recovery-vs-cos}. Right: against the concept-rate AUC of the transferred vector alone, i.e.\ the numerator of recovery.}
\label{tab:geometry-correlations}
\vspace{10pt}
\begin{tabular}{llccc|cc}
\toprule
 & & \multicolumn{3}{c|}{vs.\ recovery} & \multicolumn{2}{c}{vs.\ AUC of transferred vector} \\
Model & Direction & $\rho$ & $r$ & $p$ & $\rho$ & $p$ \\
\midrule
UniAR  & \qTI & $+0.76$ & $+0.58$ & $<0.001$ & $+0.80$ & $<0.001$ \\
UniAR  & \qIT & $+0.17$ & $+0.19$ & $0.46$ & $+0.17$ & $0.46$ \\
Emu3.5 & \qTI & $-0.27$ & $-0.19$ & $0.23$ & $+0.42$ & $0.06$ \\
Emu3.5 & \qIT & $+0.47$ & $+0.62$ & $0.03$ & $+0.04$ & $0.88$ \\
Liquid & \qTI & $-0.12$ & $-0.13$ & $0.60$ & $+0.32$ & $0.16$ \\
Liquid & \qIT & $+0.69$ & $+0.60$ & $0.001$ & $+0.35$ & $0.12$ \\
\bottomrule
\end{tabular}
\end{table}

\begin{table}[H]
\centering
\small
\caption{Anisotropy at the middle layers: mean cosine similarity between random pairs of single-token activations, raw and after centering (unit length, midpoint subtracted). Samples: $64$ tokens from each of $5$ images per concept, and every token but the first of all minimal-pair sentences.}
\label{tab:anisotropy-models}
\vspace{10pt}
\begin{tabular}{lccc|ccc}
\toprule
 & \multicolumn{3}{c|}{raw} & \multicolumn{3}{c}{unit length, centered} \\
Model & text--text & image--image & image--text & text--text & image--image & image--text \\
\midrule
UniAR  & $0.75$ & $0.74$ & $0.65$ & $0.17$ & $0.16$ & $-0.17$ \\
Emu3.5 & $0.36$ & $0.45$ & $0.17$ & $0.15$ & $0.18$ & $-0.16$ \\
Liquid & $0.51$ & $0.50$ & $0.42$ & $0.09$ & $0.07$ & $-0.08$ \\
\bottomrule
\end{tabular}
\end{table}

\begin{table}[H]
\centering
\small
\setlength{\tabcolsep}{4pt}
\caption{Same concept pole vs.\ same direction at the middle layers, per concept and model. \emph{pole cos}: cosine, after centering, between the positive-pole centroid computed from images and from sentences. \emph{ratio}: distance from the positive-image centroid to the positive-text centroid, divided by its distance to the negative-image centroid (independent of centering). \emph{dir.\ cos}: cosine between the text and image steering vectors, with $z$ = cosine divided by the baseline standard deviation.}
\label{tab:objects-models}
\vspace{10pt}
\begin{tabular}{l|cc c|cc c|cc c}
\toprule
 & \multicolumn{3}{c|}{UniAR} & \multicolumn{3}{c|}{Emu3.5} & \multicolumn{3}{c}{Liquid} \\
Concept & pole cos & ratio & dir.\ cos ($z$) & pole cos & ratio & dir.\ cos ($z$) & pole cos & ratio & dir.\ cos ($z$) \\
\midrule
emotion    & $-0.34$ & $53\times$ & $0.47$ (30) & $-0.62$ & $203\times$ & $0.09$ (6) & $-0.33$ & $210\times$ & $0.02$ (1) \\
age        & $-0.49$ & $25\times$ & $0.31$ (20) & $-0.65$ & $33\times$ & $0.07$ (5) & $-0.41$ & $25\times$ & $0.05$ (3) \\
cleanness  & $-0.46$ & $10\times$ & $0.20$ (13) & $-0.65$ & $12\times$ & $0.01$ (1) & $-0.36$ & $9\times$ & $0.08$ (5) \\
chaos      & $-0.42$ & $12\times$ & $0.14$ (9) & $-0.61$ & $23\times$ & $0.00$ (0) & $-0.39$ & $17\times$ & $0.07$ (4) \\
size       & $-0.59$ & $15\times$ & $-0.02$ ($-1$) & $-0.76$ & $25\times$ & $-0.02$ ($-1$) & $-0.59$ & $11\times$ & $-0.01$ ($-1$) \\
depth      & $-0.52$ & $7\times$ & $0.03$ (2) & $-0.64$ & $12\times$ & $-0.01$ ($-1$) & $-0.52$ & $6\times$ & $0.08$ (4) \\
spatial    & $-0.44$ & $40\times$ & $-0.02$ ($-2$) & $-0.54$ & $57\times$ & $0.01$ (1) & $-0.34$ & $48\times$ & $0.03$ (2) \\
color      & $-0.31$ & $17\times$ & $0.60$ (39) & $-0.63$ & $15\times$ & $0.19$ (14) & $-0.31$ & $9\times$ & $0.36$ (20) \\
\bottomrule
\end{tabular}
\end{table}


\section{Concepts unique to one modality}
\label{app:native-examples}
This appendix provides details on \Cref{sec:qualitative}: all modality-unique concepts we could conceive, and examples for the concepts that were successful.

\subsection{All attempted concepts}
\label{app:tried}
A candidate has to be native to one modality and still expressible as two poles of one
dimension. Text pairs are $100$ sentences from one template with the attribute at the final
token; language identity instead uses $100$ parallel translations, mean-pooled, since a
language is a property of a whole sentence. Image pairs are built programmatically where the
property is an image operation (a directional blur for motion blur, a reduction to
$48\times48$ pixels for resolution) and otherwise as two edits of one generated image (camera
height). We check that a direction steers its own modality before applying it, unchanged, to
the other, and we report each concept on prompts where the property can appear at all: steering
formality while the model describes a bowl of fruit changes little, while on a message to a
friend the same vector moves contractions from $49$ to $0$ over twenty prompts. Where free
generation shows nothing, we let the model complete a sentence whose next word must pick a
pole, e.g.\ \emph{``The shot was taken from \underline{\hspace{6mm}}''}, for which it says
\emph{below} $1\%$, $15\%$ and $89\%$ of the time at $\alpha=-48,0,+48$, while eight random
directions of the same size move the choice by $-0.17\pm0.59$ nats against $+4.50$ here.
\Cref{tab:tried} lists the concepts we built pairs for. Two failed at that stage: for render
vs.\ photograph the editor either changed nothing or simply desaturated the image, and for
first- vs.\ third-person viewpoint it could not turn an outside photograph into a
participant's view (one usable pair in twelve).
Visual concepts that represent a mechanical operation or a
small camera move can be used as minimal pairs. 
But concepts that require the editing model to reimagine the whole scene do not.

\begin{table}[H]\centering\footnotesize
\caption{The concepts we built minimal pairs for, and what came of each.}
\label{tab:tried}
\vspace{10pt}
\begin{tabular}{@{}p{2.7cm}p{2.3cm}p{3.1cm}p{3.9cm}@{}}
\toprule
Concept & Poles & Pairs & Outcome \\
\midrule
\multicolumn{4}{@{}l}{\emph{Language-native}}\\
formality & colloquial / formal & text template & works in both modalities \\
language identity & English / French & parallel translations & works in both modalities \\
language identity & English / Chinese & parallel translations & works in text; in images mostly adds lettering \\
\midrule
\multicolumn{4}{@{}l}{\emph{Vision-native}}\\
motion blur & sharp / blurred & directional blur of one image & works in both modalities \\
resolution & high / low & $48{\times}48$ pixelation of one image & works in both modalities \\
camera height & above / below & two edits of one baseline & steers images; text only in forced choice \\
\midrule
\multicolumn{4}{@{}l}{\emph{Tried and dropped}}\\
render vs.\ photo & rendered / photo & editor & pairs failed \\
viewpoint & first / third person & editor & pairs failed \\
\bottomrule
\end{tabular}
\end{table}

\FloatBarrier
\subsection{Examples for the concepts that worked}
\label{app:worked}
In every figure the middle column is the unsteered baseline and the columns are steering
strength relative to the largest value of that sweep, as in \Cref{fig:qual}; each row is one
prompt, listed in the caption.

\paragraph{Formality.} Asked to \emph{``write an email to a friend inviting them to dinner''},
the model opens with \emph{``Hey [Friend's Name], I'm so excited\ldots''} at $\alpha=-16$ and
with \emph{``Dear [Friend's Name], \ldots I am writing to extend a warm invitation\ldots''} at
$\alpha=+16$; the formal pole drops contractions and adds letter conventions. At high strength
it also changes content: an apology for a noisy party comes to claim ``the noise was not my
fault''.
\begin{figure}[H]\centering
\includegraphics[width=\textwidth]{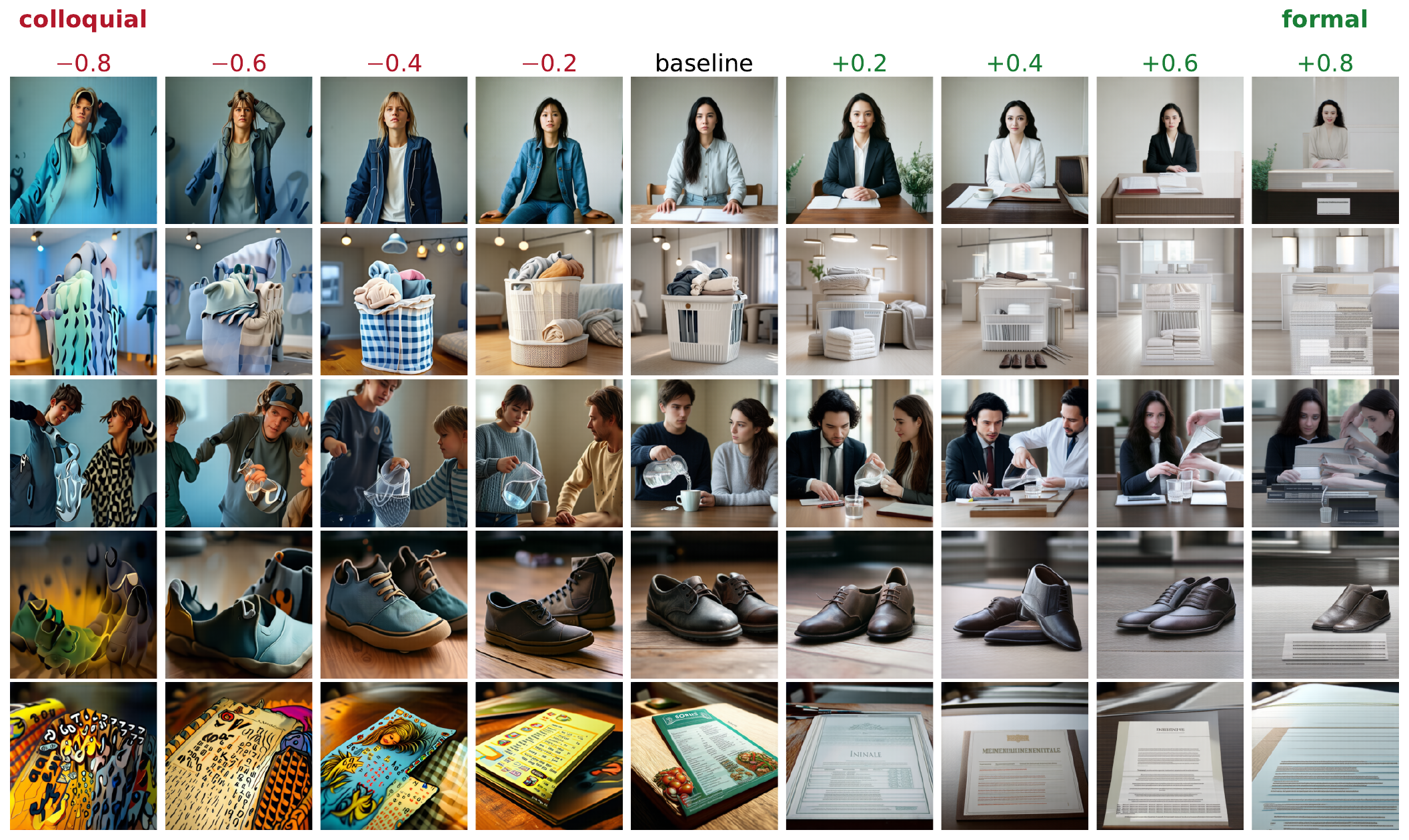}
\caption{\textbf{Formality, text $\to$ image.} Rows: a person at a desk, a basket of laundry,
one person pouring water for another, a pair of shoes, a menu. None of the prompts mentions
clothing or register. Toward formal: suits, documents, neatly stacked laundry, dress shoes, a
plain typeset menu. Toward colloquial the scenes become warmer and more cluttered, and beyond
$-0.6$ they start to look like illustrations; beyond $+0.6$ they wash out.}
\label{fig:native-formality}
\end{figure}

\FloatBarrier
\paragraph{Language identity (English--French).} Since an English prompt is already answered in
English, we prompt in German: \emph{``Beschreibe eine kleine B\"ackerei an einer
Stra\ss enecke''} is answered in German unsteered, in English at $\alpha=-16$, and in French
from $\alpha=+16$ (\emph{``Le soleil se l\`eve lentement derri\`ere les toits\ldots''}). In
images the objects themselves take a French form, and any writing turns into French-looking but
non-existent words. The direction supplies cultural convention only where the prompt leaves it
open: scenes explicitly placed in East Asia stay East Asian at every strength.
\begin{figure}[H]\centering
\includegraphics[width=\textwidth]{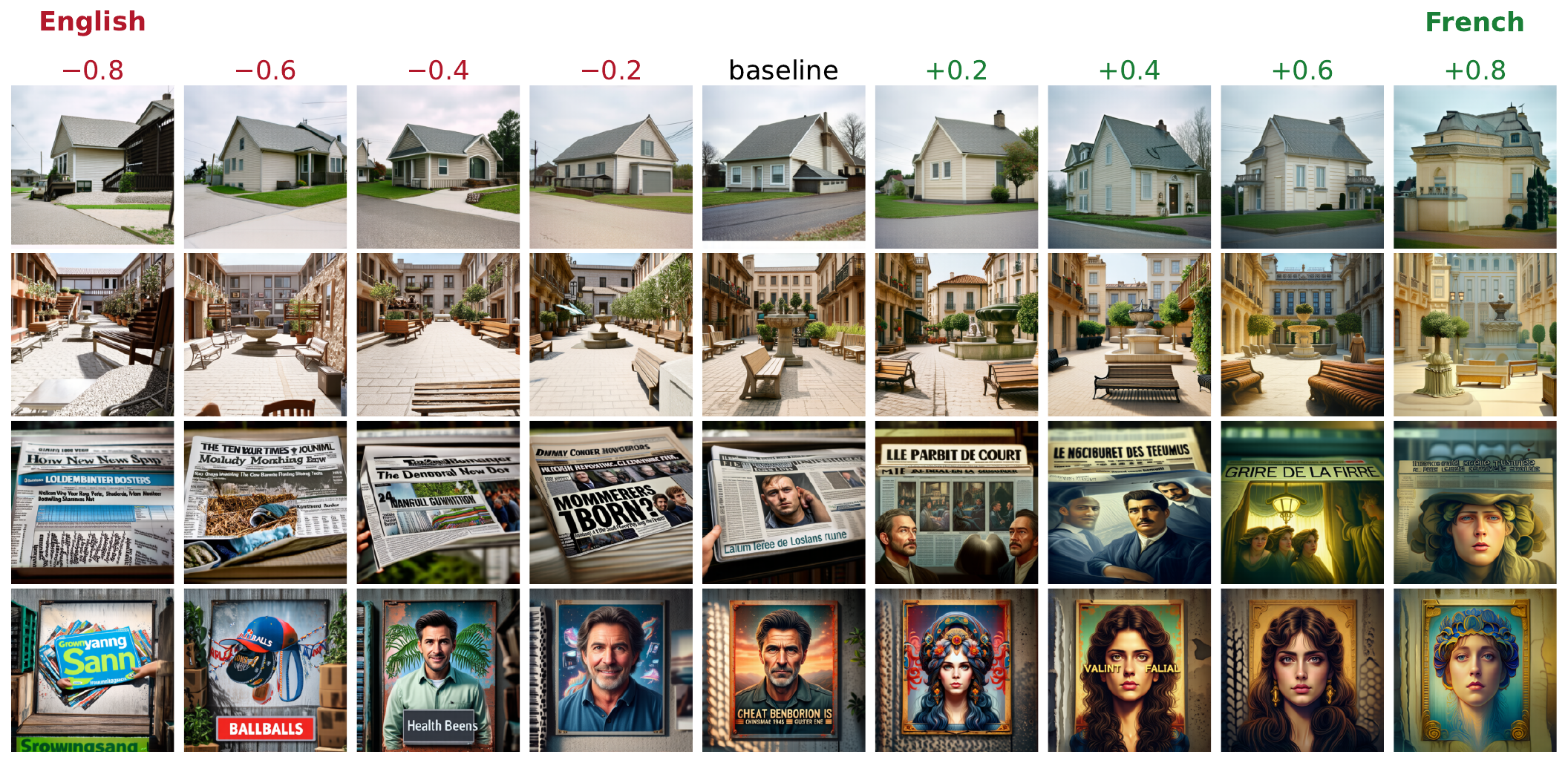}
\caption{\textbf{English vs.\ French, text $\to$ image.} Rows: a house, a town square, the
front page of a newspaper, a poster. Toward French: a mansard roof, a Beaux-Arts square,
mastheads and posters in French-looking but invented words. Toward English: vinyl siding and a
garage, plain paving, English headlines and slogans.}
\label{fig:native-langid}
\end{figure}
\begin{figure}[H]\centering
\includegraphics[width=\textwidth]{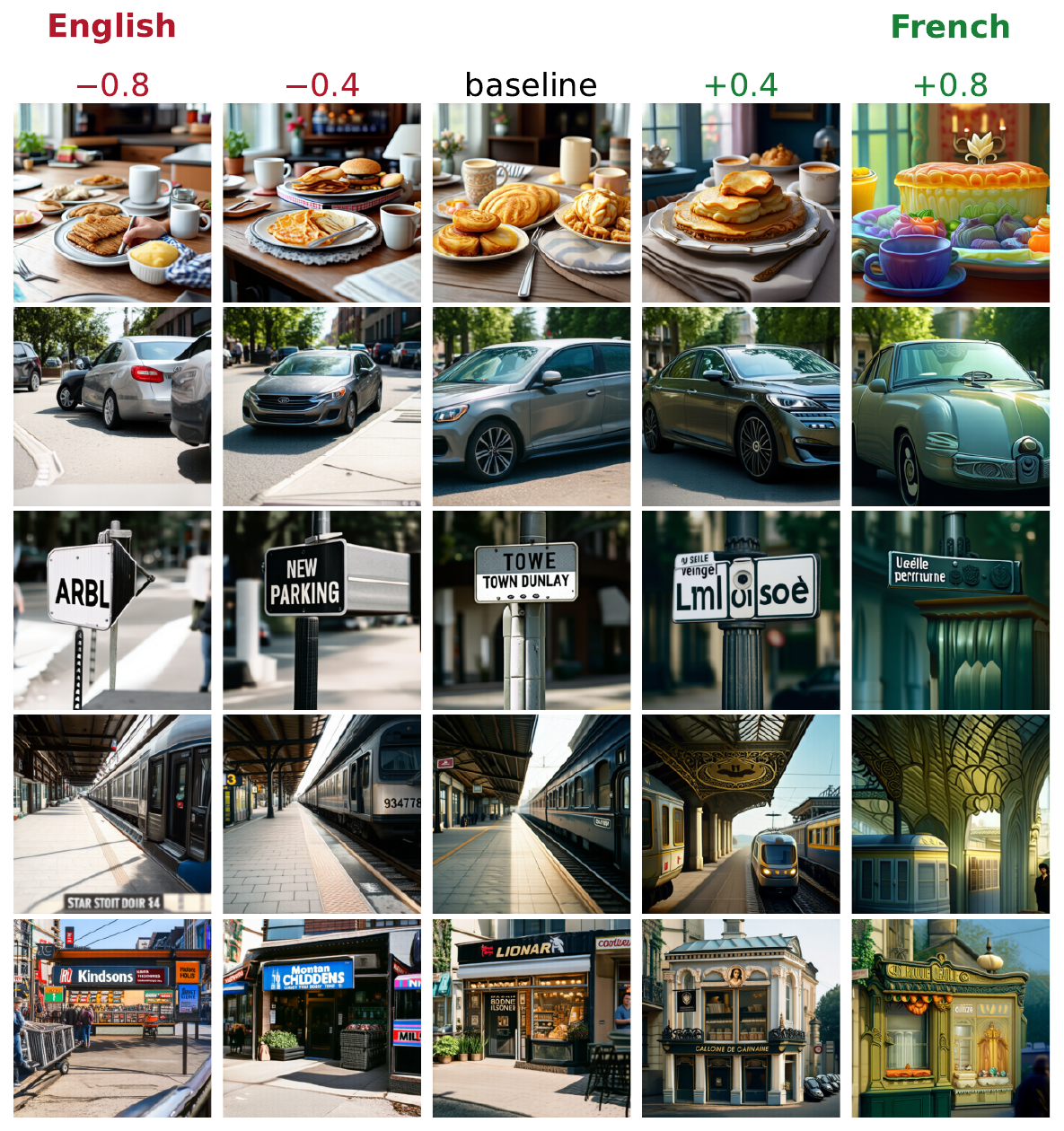}
\caption{\textbf{The same direction on everyday objects.} Rows: a breakfast, a parked car, a
street sign, a station platform, a small shop. At $+0.4$ the objects are French while the image
is still photographic (pastries, a vintage car, an enamel street sign, an iron station canopy,
a Haussmann shopfront); at $+0.8$ the images turn golden and painterly.}
\label{fig:native-langid-diverse}
\end{figure}

\FloatBarrier
\paragraph{Camera height.} The direction steers images cleanly in both directions but leaves
free text almost untouched. We probed it with two prompts, reading the next-token distribution
at layers $25$--$28$. Completing \emph{``The shot was taken from \underline{\hspace{4mm}}''},
steering toward a low camera raises $P(\text{\emph{below}})$ from $3.7\%$ to $12.8\%$ and lowers
$P(\text{\emph{above}})$ from $21.1\%$ to $4.7\%$ ($\alpha=+32$), but the most likely token remains
a function word at every strength. Asked directly, \emph{``Was the shot taken from above or
below? Answer:''}, the model answers \emph{above} unsteered ($45\%$ against $12\%$) and still
answers \emph{above} at every strength, with $P(\text{\emph{below}})$ peaking at $16\%$; swapping
the order of the two words in the question moves the answer more than the steering does.
\begin{figure}[H]\centering
\includegraphics[width=\textwidth]{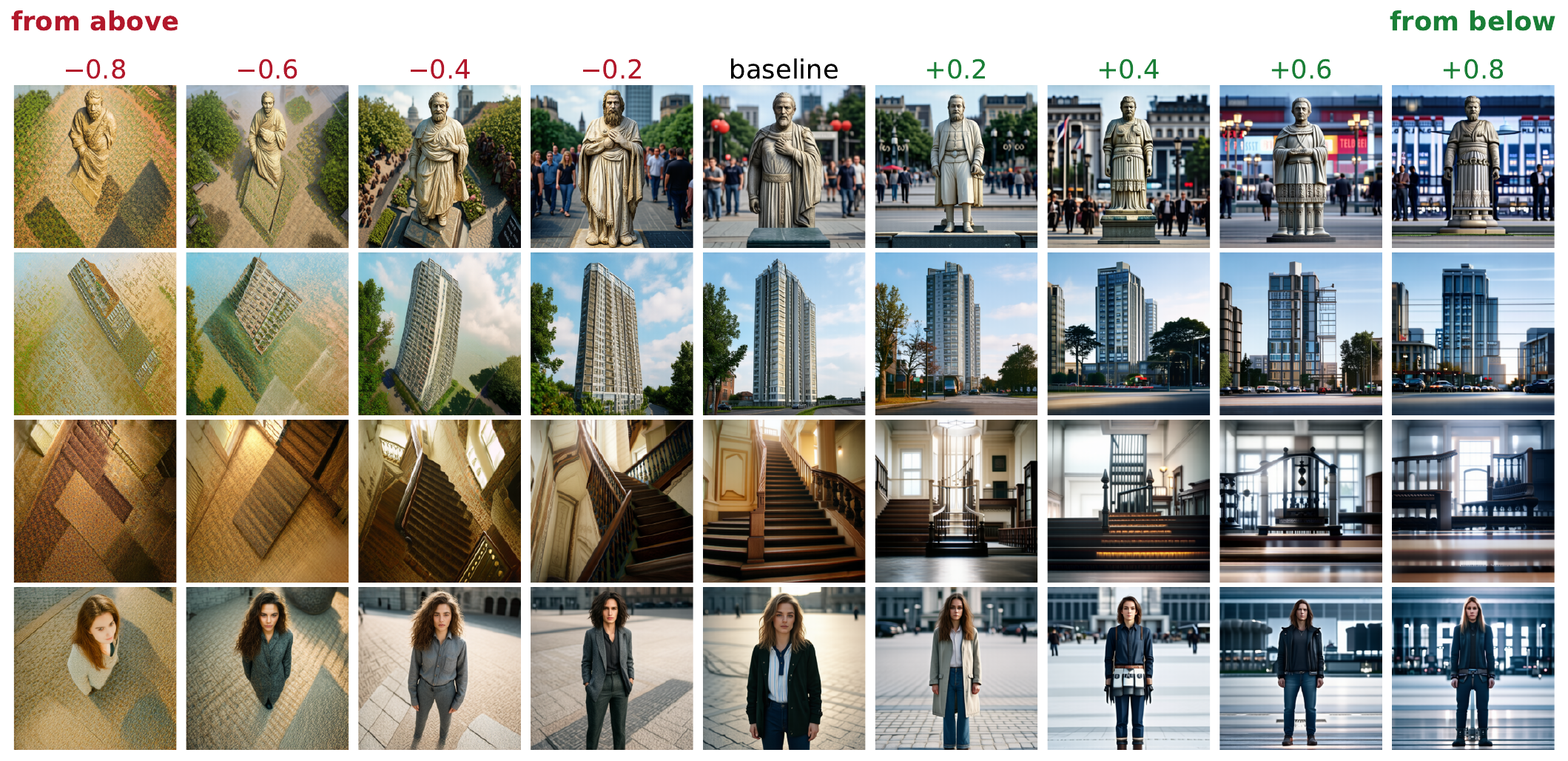}
\caption{\textbf{Camera height, image $\to$ image.} Rows: a statue, an apartment tower, a
staircase, a person. Toward one pole the camera rises until it looks straight down on the
subject; toward the other it drops to near ground level. The subject stays the same.}
\label{fig:native-ch}
\end{figure}

\FloatBarrier
\paragraph{Motion blur.} In free text, steering toward sharp makes descriptions calmer
(\emph{serene}, \emph{calm}: $3$ occurrences unsteered, $11$ at $\alpha=-32$) and steering
toward blurred adds movement and speed words, and finally \emph{blur} itself: asked for the
view from a car, the model writes ``a blur of motion and colors as the vehicle moves at high
speed''. In forced choice it moves ``the runner looks \underline{\hspace{6mm}}'' from $6\%$ to
$99\%$ \emph{blurry}.
\begin{figure}[H]\centering
\includegraphics[width=\textwidth]{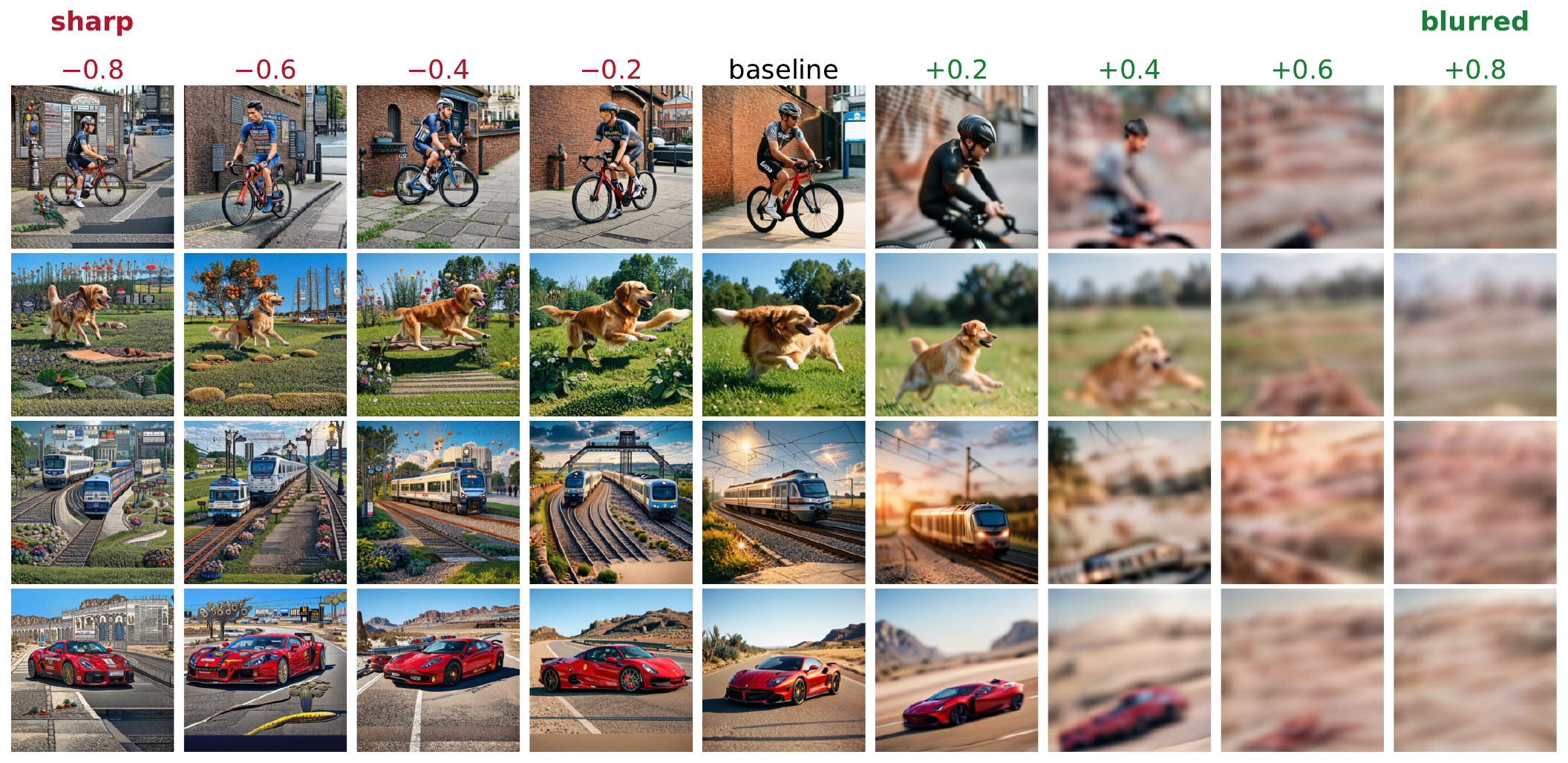}
\caption{\textbf{Motion blur, image $\to$ image.} Rows: a cyclist, a dog, a train, a sports
car. Toward blurred, the background first streaks as in a panning shot while the subject stays
readable, then the whole frame blurs. Toward sharp, the scene fills with extra detail.}
\label{fig:native-mb}
\end{figure}

\FloatBarrier
\paragraph{Resolution.} Steering toward low resolution makes descriptions short and vague:
\emph{``I'm sitting in a window with a blurry view of the city. \ldots I'm not sure what I'm
doing''} ($36$ words) against $65$ words of specific detail unsteered; the words \emph{blurry}
and \emph{pixelated} appear only toward low resolution. The shortening is specific to this
direction: other directions of the same size, including random ones, leave answers at
$57$--$73$ words.
\begin{table}[H]\centering\footnotesize
\begin{tabular}{@{}p{2.4cm}p{5.2cm}p{5.2cm}@{}}
\toprule
\textbf{must include} & \textbf{unsteered} & \textbf{toward low resolution} \\
\midrule
a street with shops & \emph{``I strolled down the bustling street, where the scent of fresh bread and coffee filled the air. I passed by a cozy caf\'e, its inviting aroma drawing me in. [\ldots]''} & \emph{``I'm walking down a busy street in the city. I see a shop with a bright red sign.''} \\
\addlinespace[2pt]
a view from a window & \emph{``I wake up to the soft hum of the city, the distant clatter of traffic, and the faint scent of coffee and new beginnings. [\ldots]''} & \emph{``I'm sitting in a window with a \textbf{blurry} view of the city. It's a bright day, and I can see some buildings in the background. I'm \textbf{not sure what I'm doing}, but I'm just sitting there.''} \\
\addlinespace[2pt]
a bridge over the river & \emph{``I wake up to the soft hum of the city, the distant sound of traffic and the gentle lapping of water. As I step out onto the bridge, [\ldots]''} & \emph{``I was walking through the city, and I saw a bridge over the river. I was walking on the bridge, and I was looking at the city.''} \\
\addlinespace[2pt]
a hotel lobby & \emph{``I wake up in the hotel lobby, the soft hum of the alarm clock filling the air. The lobby is quiet, with a few guests lingering in the background. [\ldots]''} & \emph{``I'm in a hotel lobby, and I'm trying to find something. I see a \textbf{pixelated} object that looks like a \textbf{pixelated} character. I \textbf{can't tell what it is}, but I'm trying to understand it.''} \\
\addlinespace[2pt]
a construction site & \emph{``I wake up to the soft hum of the city, the distant clatter of traffic, and the faint scent of coffee. Today, I'm heading to the construction site, where the early morning light paints the steel and concrete in hues of gold and silver. [\ldots]''} & \emph{``I'm working on a project at a construction site. I'm trying to fix something, but I \textbf{can't see what I'm fixing}.''} \\
\addlinespace[2pt]
a small alley & \emph{``I strolled through the city, the air thick with the scent of coffee and the distant hum of traffic. As I walked down a narrow alley, the walls on either side were lined with small shops and cafes. [\ldots]''} & \emph{``I was walking through a small alley when I saw a person who \textbf{looked like a ghost}.''} \\
\addlinespace[2pt]
\bottomrule
\end{tabular}
\caption{Six examples of applying the high--low resolution steering vector to text generation,
for the prompt \emph{``Describe your day in the city. Your description must include [X]''}
($\alpha=+32$).}
\label{tab:res-text}
\end{table}

\begin{figure}[H]\centering
\includegraphics[width=\textwidth]{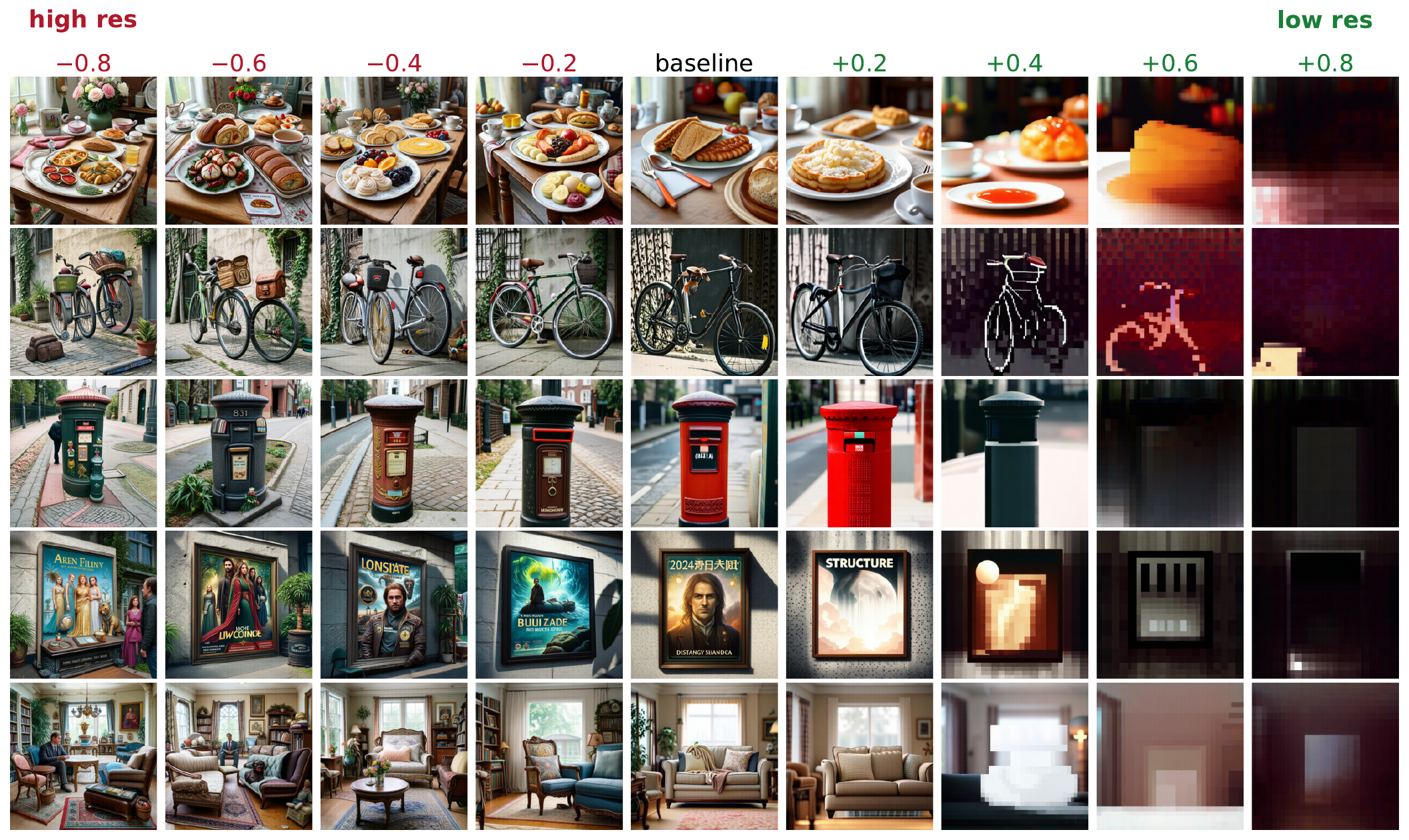}
\caption{\textbf{Resolution, image $\to$ image.} Rows: a breakfast, a bicycle, a postbox, a
poster, a living room. Toward high resolution scenes gain detail and clutter; toward low
resolution they first become simpler and emptier, then turn into pixel blocks, and collapse
at $+0.8$.}
\label{fig:native-res}
\end{figure}

\FloatBarrier

\section{Averaging text and image steering vectors}
\label{app:averaging}

Here, we report a negative finding: averaging the steering vectors from the two modalities
(which is equivalent to applying both at the same time) does not seem to yield any
straightforward \textit{quantitative} improvements. We did not qualitatively inspect whether
averaging has any interesting effects.

For each concept we form the per-layer average
$\hat v(\lambda)=\mathrm{normalize}\big(\lambda\,\hat v_{\text{txt}}+(1-\lambda)\,\hat v_{\text{img}}\big)$
with $\lambda\in\{0.25,0.5,0.75\}$, so that $\lambda\!=\!1$ and $\lambda\!=\!0$ recover the
single-modality vectors, and additionally a baseline that averages $\hat v_{\text{txt}}$ with
a random direction of matched norm. We steer text generation with each vector at the early,
middle, and late layer ranges over the full strength grid, and score every output with the
concept-rate judge against the unsteered baseline, summarized as the per-pole maximum and
the range-normalized AUC.

No average improves over the better single-modality vector: at the per-pole maximum the best
average exactly ties the text vector, and differences elsewhere are within judge noise
(\Cref{tab:averaging}); the same holds at the other layer ranges and on the image target,
where rates saturate for most vectors.

\begin{table}[H]\centering
\small
\caption{Averaging the two modalities' steering vectors, text target, middle layer range:
concept rate summarized as per-pole maximum and range-normalized AUC over
$\alpha\in[-60,60]$, averaged over the six transfer concepts. No average improves over the
better single-modality vector beyond judge noise.}
\label{tab:averaging}
\vspace{10pt}
\begin{tabular}{lcc}
\toprule
Steering vector & Per-pole max & Norm.\ AUC \\
\midrule
$\hat v_{\text{txt}}$ ($\lambda\!=\!1$)            & 0.76 & 0.56 \\
$\hat v_{\text{img}}$ ($\lambda\!=\!0$)            & 0.70 & 0.47 \\
Average, $\lambda\!=\!0.75$                        & 0.76 & 0.58 \\
Average, $\lambda\!=\!0.50$                        & 0.73 & 0.55 \\
Average, $\lambda\!=\!0.25$                        & 0.71 & 0.49 \\
Random-direction control                           & 0.72 & 0.50 \\
\bottomrule
\end{tabular}
\end{table}

\clearpage





\clearpage
\section{Behind the scenes}
\label{app:behind-the-scenes}

The goal of this section is to show the ``behind the scenes'' beyond the final polished paper:
Why we started working on this, all the detours and failures, and what lessons we learned.
It is meant to make science more fun, transparent and accessible to newcomers\footnote{\href{https://bennokrojer.com/behind-the-scenes.html}{bennokrojer.com/behind-the-scenes}}.
We show both DZ's side of the story, i.e., the first author, and also BK's side as the last author.

\subsection{Junior perspective}

\textbf{How it started.} This project was my first real experience doing research. It grew out of a personal interest in vision-language models (VLMs), following a natural language understanding course project in which we studied their behaviour on spatial reasoning tasks. Over time, I found myself reading more about new ideas for visual reasoning, reasoning in general, and various other topics that caught my attention. Eventually, I came across mechanistic interpretability and began exploring model internals. I was particularly fascinated by the residual stream and the idea that we could look beyond a model's outputs to investigate how information is represented and processed within it. At that point, I had no research question in mind, just an interest in a low-level understanding of these models.

\textbf{Interpretability looks cool.} I started digging deeper into interpretability, exploring work on model circuits, activation patching, causal tracing, superposition, and sparse autoencoders. I spent quite some time reading on these ideas and began exchanging thoughts and discussing possible project ideas with Benno, who would supervise the research. We explored an initial direction with activation patching, but it never quite developed into something we were satisfied with. Benno then suggested activation steering, which offered a simple, practical, and I would argue, elegant way to intervene in a model's internal representations and observe how its behavior changed.

\textbf{Humble beginnings.} Our first candidate was Meta's Chameleon, which seemed almost ideal on paper: a simple, unified architecture with decent image generation capabilities. Unfortunately, its image generation components were not publicly released. We then considered alternatives such as Anole, but the additional training required was beyond our ``commitment budget''. Eventually, we settled on Janus, and later UniToken, which finally allowed us to start experimenting. Early results were fairly encouraging for a few concepts, which became our go-to smoke tests for subsequent experiments. However, image generation was a strong limitation, with distorted faces and various other visual artifacts. These models were useful for testing our ideas, but their image generation quality made it hard to rigorously draw meaningful conclusions or properly support our main research question.

\textbf{Tough luck.} Alongside our main model, UniAR, we also experimented with Bagel, arguably one of the strongest and more well-known open-source unified multimodal models. Its image generation capabilities were impressive (e.g., very pretty images), but its diffusion-based image generation introduced a lot of complexity. Additionally, although its pre-training was strongly unified, we weren't entirely sure how unified the model itself really was. This led to many discussions about whether it was appropriate for addressing our main research question.

\textbf{The right fit.} Model selection remained one of the biggest challenges throughout the project. Although we wanted to answer broader questions about how vision and language interact within multimodal models, we were quite particular about the models we used. We specifically wanted unified models in which both modalities are processed within a shared architecture rather than studying cross-modal interactions through separate components. However, finding a model that was sufficiently unified while also having strong image generation capabilities turned out to be surprisingly difficult.

One day, while scrolling through Rednote (a social media platform), I came across a post showing the UniAR paper from the Qwen team and Fudan University. After discussing it with Benno, we quickly moved on to our usual smoke tests, and the results looked very interesting. Surprisingly, even some concepts that hadn't worked at all with previous models, such as depth near/far, showed encouraging results. The image generation quality was, overall, decently impressive (the generated images were very pretty). After all of it, UniAR finally seemed to be what we had been looking for.

\textbf{The bigger picture.} Throughout the project, my discussions with Benno, particularly his interest in more fundamental scientific questions, gradually changed how I looked at our work and inspired me to be more curious. Initially, it felt like a series of experiments on manipulating model behavior. It then became: how do vision and language interact, to what extent do they share a common understanding of the world, and more. After Tamar and Nishant joined the project, I also had the chance to witness great minds debate novel ideas, navigate through ambiguity in a strong, systematic way, and develop creative yet rigorous ways to test new hypotheses.

I hope our work can serve as a useful starting point for studying cross-modal behaviour. Beyond demonstrating where concrete cross-modal steering methods succeed or fail, I hope our findings can help others better understand and study the relationship between visual and linguistic representations. There are still so many abstract ideas, but I think that is precisely what makes this direction worth exploring!

\subsection{Senior perspective}
This project was unique because it was DZ's first time doing research and BK's first time fully supervising an undergraduate end-to-end, from ideation to our final submission.
BK had mentored undergraduates before but in those cases the student either had already some experience and direction in mind or it did not end up as a final conference submission.
So, this mentoring from a ``blank slate'' to a full conference submission was a proud moment!

\paragraph{The project itself.}
When DZ approached me for an ``independent study'' at McGill, the ideas were still broadly around visual concepts in models.
There was a lot of motivation and curiosity but a lot of scoping to be done.
So we brainstormed for a few weeks and, while I don't remember how exactly, eventually we stumbled upon the idea of transferring steering vectors across modalities.
I thought this would be very elegant and novel to study, and probably under-explored.
More broadly I had been waiting for a chance to do interpretability on unified multimodal models, as opposed to the more mainstream VLMs.
So the project properly took this shape around late January 2026, and then as DZ describes above, it was a many-months journey of finding the right models to study.
The lesson here is that for a lot of interesting phenomena, one has to wait for the right moment, such as model capabilities or architectures to mature.
In our case models had to get a bit better at generating images perhaps, especially fully autoregressive unified models.
UniAR is a relatively unknown and new model but came at the right time for us \citep{peng2026unified}.

\paragraph{Thoughts on modality unification.}
It is unclear if we are close to reaching the grand goal that made researchers develop unified multimodal models in the first place.
Intuitively, there should be lots of synergies and overlaps between visual and textual capabilities.
Therefore, the goal is broadly that we want a single model that can generate anything from the same underlying representation space.
However from the capabilities angle, these models have not really replaced mainstream LLaVA-style VLMs or diffusion-based text-to-image models.
And from an analysis angle, there has been some early evidence that these unified models might not actually be particularly unified internally \citep{cite_narrow_gate, gao2026text}.
One also has to simply acknowledge that the design space here is much vaster and trickier than for pure text or image generation \citep{tong2026beyond}.
So while analysis work on earlier unified models might have shown mostly negative results, newer models, with years of engineering insights baked into them, might be much more unified.
But for the next couple years, I am still not too optimistic about this line of work.
For example, interpretability studies on unified models might still be considered less mainstream or impactful.
But long-term there is still optimism that these models will become the mainstream paradigm, since they follow the spirit that has dominated AI development in the last decade:
Instead of designing different models for different tasks or domains, build one general model.
This will then open up exciting and difficult challenges for interpretability and cognitive science, such as much more interesting questions about acquisition during pre-training.
For someone considering working on them from this analysis angle, I would recommend to go for it but to be aware that this field might not be the most impactful or trendy for a while.

\paragraph{Reflections on mentoring, becoming a senior researcher and beyond.}
Some of this project was completed at the end of BK's PhD and some during the newly started postdoc.
It therefore represents very well the transition from being a hands-on PhD student to a more senior role of mentorship and supervision.
Such a transition is not easy and requires very different skills than those needed for hands-on leading a project.
So just like DZ, BK also learned a lot.
Some reflections on the different kind of skills this senior role brings forth, compared to the usual skills acquired during the first years of a PhD:
\begin{enumerate}
    \item After being used to having control over every aspect of the project as a PhD student, it will not be easy to let go of such control and trust the student and the process.
    \item On the other hand, it is equally important to know when to take over from the student. This is especially true for writing where it is unrealistic to expect an undergraduate to know all the unspoken rules and nuances of a conference submission paper. So BK ended up framing or writing large chunks of the paper based on DZ's initial results and drafts, and with help from the other two senior authors.
    \item It is also not an easy balance what kind of Claude Code habits to recommend, such that the student still learns by themselves mostly.
    \item There is not a single best way of mentoring so adding a few other more senior roles was very helpful, both for me to refine my ideas with another senior person, and for DZ to witness different approaches to science, project habits and paper writing. It is a very useful skill to know who could be good additions to a project, especially when there is no actual PI on the project.
\end{enumerate}

Even when you are not yet close to finishing your PhD, I highly recommend mentoring undergrads.
But I also recommend being realistic about it.
Do not pour your hopes into the project immediately, expecting them to be like a mini-PhD student from day one.
Instead, the more you see evidence of scientific curiosity, self-initiative and long-term commitment, the more you can gradually increase your expectations and time investment.
The beautiful part of such experiences is how it re-connects you with the younger version of yourself when you entered research, perhaps naive but also full of excitement and energy.
It teaches empathy and lets you see the process of science from a birds-eye-view: a never-ending cycle of standing on the shoulders of giants, of continuous mentoring and nurturing a spirit of exploration.

\end{document}